\documentclass[letterpaper]{article}
\usepackage{uai}
\usepackage[margin=1in,headsep=0.13in,headheight=11pt]{geometry}

\usepackage{caption}
\usepackage{times}
\usepackage{graphicx} 
\usepackage{subfigure} 
\usepackage{dblfloatfix} 

\usepackage{natbib}

\usepackage{algorithm}
\usepackage{algorithmic}

\usepackage{amsmath}
\usepackage{amssymb}
\usepackage{bbm}
\usepackage[usenames, dvipsnames]{color}
\newcommand\independent{\protect\mathpalette{\protect\independenT}{\perp}}
\def\independenT#1#2{\mathrel{\rlap{$#1#2$}\mkern2mu{#1#2}}}
\newcommand{\estimates}{\overset{\scriptscriptstyle\wedge}{=}}
\newcommand{\indicator}[1]{\mathbbm{1}[[#1]]}
\usepackage{textcomp}
\usepackage{enumerate}
\newcommand\numberthis{\addtocounter{equation}{1}\tag{\theequation}}
\DeclareMathOperator*{\argmin}{\arg\!\min}
\newtheorem{theorem}{Theorem}

\newtheorem{lemma}[theorem]{Lemma}
\usepackage{booktabs}

\usepackage{lastpage}
\usepackage{newfloat}
\DeclareFloatingEnvironment[name={Figure B}]{suppfigure}
\DeclareFloatingEnvironment[name={Table C}]{supptable}

\makeatletter
\newcommand{\customlabel}[2]{%
   \protected@write \@auxout {}{\string \newlabel {#1}{{#2}{\thepage}{#2}{#1}{}} }%
   \hypertarget{#1}{#2}
}
\makeatother

\def\abovestrut#1{\rule[0in]{0in}{#1}\ignorespaces}
\def\belowstrut#1{\rule[-#1]{0in}{#1}\ignorespaces}



\usepackage{perpage} 
\MakePerPage{footnote} 

\title{Learning with Confident Examples: \\ Rank Pruning for Robust Classification with Noisy Labels}

\author{\textbf{Curtis G. Northcutt},  \thanks{\  Equal Contribution} \enskip \quad \textbf{Tailin Wu}, \footnotemark[1] \enskip  \quad \textbf{Isaac L. Chuang} \\
Massachusetts Institute of Technology\\
Cambridge, MA 02139 \\
\{cgn, tailin, ichuang\}@mit.edu \\
}

\usepackage[utf8]{inputenc}
\usepackage[english]{babel}
\usepackage[colorlinks=true,allcolors=blue]{hyperref}
\usepackage{fancyhdr}
 
\begin{document}

\maketitle

\begin{abstract}
$\tilde{P}\tilde{N}$ learning is the problem of binary classification when training examples may be mislabeled (flipped) uniformly with noise rate $\rho_1$ for positive examples and $\rho_0$ for negative examples. We propose Rank Pruning (RP) to solve $\tilde{P}\tilde{N}$ learning and the open problem of estimating the noise rates, i.e. the fraction of wrong positive and negative labels. 
Unlike prior solutions, RP is time-efficient and general, requiring $\mathcal{O}(T)$ for any unrestricted choice of probabilistic classifier with $T$ fitting time. We prove RP has consistent noise estimation and equivalent expected risk as learning with uncorrupted labels in ideal conditions, and derive closed-form solutions when conditions are non-ideal. RP achieves state-of-the-art noise estimation and F1, error, and AUC-PR for both MNIST and CIFAR datasets, regardless of the amount of noise and performs similarly impressively when a large portion of training examples are noise drawn from a third distribution. To highlight, RP with a CNN classifier can predict if an MNIST digit is a \emph{one} or \emph{not} with only $0.25\%$ error, and $0.46\%$ error across all digits, even when 50\% of positive examples are mislabeled and 50\% of observed positive labels are mislabeled negative examples.

\end{abstract}

\section{Introduction}
\label{sec:intro}

Consider a student with no knowledge of animals tasked with learning to classify whether a picture contains a dog. A teacher shows the student example pictures of lone four-legged animals, stating whether the image contains a dog or not. Unfortunately, the teacher may often make mistakes, asymmetrically, with a significantly large false positive rate, $\rho_1 \in [0,1]$, and significantly large false negative rate, $\rho_0 \in [0,1]$. The teacher may also include ``white noise" images with a uniformly random label. This information is unknown to the student, who only knows of the images and corrupted labels, but suspects that the teacher may make mistakes. Can the student (1) estimate the mistake rates, $\rho_1$ and $\rho_0$, (2) learn to classify pictures with dogs accurately, and (3) do so efficiently (e.g. less than an hour for 50 images)? This allegory clarifies the challenges of $\tilde{P}\tilde{N}$ learning for any classifier trained with corrupted labels, perhaps with intermixed noise examples. We elect the notation $\tilde{P}\tilde{N}$ to emphasize that both the positive and negative sets may contain mislabeled examples, reserving $P$ and $N$ for uncorrupted sets.

This example illustrates a fundamental reliance of supervised learning on training labels \citep{Michalski:1986:MLA:21934}. Traditional learning performance degrades monotonically with label noise \citep{Aha1991, Nettleton2010}, necessitating semi-supervised approaches \citep{Blanchard:2010:SND:1756006.1953028}. Examples of noisy datasets are medical \citep{raviv_heart}, human-labeled \citep{mech_turk_quality}, and sensor \citep{lane2010survey} datasets. The problem of uncovering the same classifications as if the data was not mislabeled is our fundamental goal.

Towards this goal, we introduce Rank Pruning\footnote{ Rank Pruning is open-source and available at \href{http://bit.ly/2pbtjR6}{https://github.com/cgnorthcutt/rankpruning}}, an algorithm for $\tilde{P}\tilde{N}$ learning composed of two sequential parts: (1) estimation of the asymmetric noise rates $\rho_1$ and $\rho_0$ and (2) removal of mislabeled examples prior to training. The fundamental mantra of Rank Pruning is \emph{learning with confident examples}, i.e. examples with a predicted probability of being positive \emph{near} $1$ when the label is positive or $0$ when the label is negative. If we imagine non-confident examples as a noise class, separate from the confident positive and negative classes, then their removal should unveil a subset of the uncorrupted data. 

An ancillary mantra of Rank Pruning is \emph{removal by rank} which elegantly exploits ranking without sorting. Instead of pruning non-confident examples by predicted probability, we estimate the number of mislabeled examples in each class. We then remove the $k^{th}$-most or $k^{th}$-least examples, \emph{ranked} by predicted probability, via the BFPRT algorithm \citep{Blum:1973:TBS:1739940.1740109} in $\mathcal{O}(n)$ time, where $n$ is the number of training examples. \emph{Removal by rank} mitigates sensitivity to probability estimation and exploits the reduced complexity of learning to rank over probability estimation \citep{Menon2012PredictingAP}. Together, \emph{learning with confident examples} and \emph{removal by rank} enable robustness, i.e. invariance to erroneous input deviation.


Beyond prediction, confident examples help estimate $\rho_1$ and $\rho_0$. Typical approaches require averaging predicted probabilities on a holdout set \citep{Liu:2016:CNL:2914183.2914328, Elkan:2008:LCO:1401890.1401920} tying noise estimation to the accuracy of the predicted probabilities, which in practice may be confounded by added noise or poor model selection. Instead, we estimate $\rho_1$ and $\rho_0$ as a fraction of the predicted counts of confident examples in each class, encouraging robustness for variation in probability estimation.

\subsection{Related Work}



Rank Pruning bridges framework, nomenclature, and application across $PU$ and $\tilde{P}\tilde{N}$ learning. In this section, we consider the contributions of Rank Pruning in both.

\subsubsection{\texorpdfstring{$PU$}{PU} Learning}


Positive-unlabeled ($PU$) learning is a binary classification task in which a subset of positive training examples are labeled, and the rest are unlabeled. For example, co-training \citep{Blum:1998:CLU:279943.279962, Nigam00understandingthe} with labeled and unlabeled examples can be framed as a $PU$ learning problem by assigning all unlabeled examples the label `0'. $PU$ learning methods often assume corrupted negative labels for the unlabeled examples $U$ such that $PU$ learning is $\tilde{P}\tilde{N}$ learning with no mislabeled examples in $P$, hence their naming conventions.

Early approaches to $PU$ learning modified the loss functions via weighted logistic regression \citep{lee2003PUlearning_weightedlogreg} and biased SVM \citep{Liu:2003:BTC:951949.952139} to penalize more when positive examples are predicted incorrectly. Bagging SVM \citep{Mordelet:2014:BSL:2565612.2565683} and RESVM \citep{Claesen201573} extended biased SVM to instead use an ensemble of classifiers trained by resampling $U$ (and $P$ for RESVM) to improve robustness \citep{Breiman:1996:BP:231986.231989}. RESVM claims state-of-the-art for $PU$ learning, but is impractically inefficient for large datasets because it requires optimization of five parameters and suffers from the pitfalls of SVM model selection \citep{Chapelle:1999:MSS:3009657.3009690}. \cite{Elkan:2008:LCO:1401890.1401920} introduce a formative time-efficient probabilistic approach (denoted \emph{Elk08}) for $PU$ learning that directly estimates $1 - \rho_1$ by averaging predicted probabilities of a holdout set and dividing all predicted probabilities by $1 - \rho_1$. On the SwissProt database, \emph{Elk08} was 621 times faster than biased SVM, which only requires two parameter optimization. However, \emph{Elk08} noise rate estimation is sensitive to inexact probability estimation and both RESVM and \emph{Elk08} assume $P$ = $\tilde{P}$ and do not generalize to $\tilde{P}\tilde{N}$ learning.  Rank Pruning leverages \emph{Elk08} to initialize $\rho_1$, but then re-estimates $\rho_1$ using confident examples for both robustness (RESVM) and efficiency (\emph{Elk08}).

\begin{table*}[t]
\caption{Variable definitions and descriptions for $\tilde{P}\tilde{N}$ learning and  PU learning. Related work contains a prominent author using each variable. $\rho_1$ is also referred to as \emph{contamination} in PU learning literature.}
 \vskip -0.15in
\label{t:definitions}
\begin{center}
\begin{small}
\begin{sc}

\resizebox{\textwidth}{!}{
\begin{tabular}{lcccr}
\toprule
\abovestrut{0.13in}\belowstrut{0.05in}
\textbf{Variable} & \textbf{Conditional} & \textbf{Description} & \textbf{Domain} & \textbf{Related} Work \\
\midrule
\abovestrut{0.1in}

$\rho_0$ & $P(s=1|y=0)$ & Fraction of $N$ examples mislabeled as positive & $\tilde{P}\tilde{N}$ & Liu \\
$\rho_1$ & $P(s=0|y=1)$ & Fraction of $P$ examples mislabeled as negative & $\tilde{P}\tilde{N}$, PU & Liu, Claesen \\
$\pi_0$ & $P(y=1|s=0)$ & Fraction of mislabeled examples in $\tilde{N}$ & $\tilde{P}\tilde{N}$ & Scott \\
$\pi_1$ & $P(y=0|s=1)$ & Fraction of mislabeled examples in $\tilde{P}$ & $\tilde{P}\tilde{N}$ & Scott \\
$c = 1- \rho_1$ & $P(s=1|y=1)$ & Fraction of correctly labeled $P$ if $P(y=1|s=1) = 1$ & PU & Elkan \\

\bottomrule
\end{tabular}
}
\end{sc}
\end{small}
\end{center}
 \vskip -0.25in
\end{table*}

\begin{table*}[b]
\renewcommand{\arraystretch}{.5}
\caption{Summary of state-of-the-art and selected general solutions to $\tilde{P}\tilde{N}$ and $PU$ learning.}
\vskip -.17in
\label{t:related_works}
\begin{center}
\begin{small}
\begin{sc}
\resizebox{\textwidth}{!}{
\begin{tabular}{lcccccccc}
\toprule
\abovestrut{0.05in}\belowstrut{0.05in}
\textbf{Related Work} & \textbf{Noise} & \textbf{$\tilde{P}\tilde{N}$} & \textbf{$PU$} & \textbf{Any Prob.} & \textbf{Prob Estim.} &                                     \textbf{Time}         & \textbf{Theory} & \textbf{Added} \\
                      & \textbf{Estim.} &                               &               & \textbf{Classifier} & \textbf{Robustness} &               \textbf{Efficient} & \textbf{Support} & \textbf{Noise} \\
\midrule
\abovestrut{0.1in}

\cite{Elkan:2008:LCO:1401890.1401920} & \checkmark &            & \checkmark & \checkmark &            & \checkmark & \checkmark & \\
\midrule
\cite{Claesen201573} &            &            & \checkmark &            & \checkmark &            &   &         \\
\midrule
\cite{ScottBH13} & \checkmark &            &            & \checkmark & \checkmark &            & \checkmark &\\
\midrule
\cite{NIPS2013_5073} &            & \checkmark & \checkmark & \checkmark & \checkmark & \checkmark & \checkmark &\\
\midrule
\cite{Liu:2016:CNL:2914183.2914328} &            & \checkmark & \checkmark & \checkmark &            & \checkmark & \checkmark &\\
\midrule
\midrule
\textbf{Rank Pruning} & \checkmark & \checkmark & \checkmark & \checkmark & \checkmark & \checkmark & \checkmark & \checkmark \\

\bottomrule
\end{tabular}
}
\end{sc}
\end{small}
\end{center}
\end{table*}

\subsubsection{\texorpdfstring{$\tilde{P}\tilde{N}$}{Noisy PN} Learning}

Theoretical approaches for $\tilde{P}\tilde{N}$ learning often have two steps: (1) estimate the noise rates, $\rho_1$, $\rho_0$, and (2) use $\rho_1$, $\rho_0$ for prediction. To our knowledge, Rank Pruning is the only time-efficient solution for the open problem \mbox{\citep{Liu:2016:CNL:2914183.2914328, ICML2012Yang_127}} of noise estimation.

We first consider relevant work in noise rate estimation. \cite{ScottBH13} established a lower bound method for estimating the \emph{inversed} noise rates $\pi_1$ and $\pi_0$ (defined in Table \ref{t:definitions}). However, the method can be intractable due to unbounded convergence and assumes that the positive and negative distributions are mutually irreducible. Under additional assumptions, \cite{scott2015rate} proposed a time-efficient method for noise rate estimation, but reported poor performance \cite{Liu:2016:CNL:2914183.2914328}. \cite{Liu:2016:CNL:2914183.2914328} used the minimum predicted probabilities as the noise rates, which often yields futile estimates of min = $0$. \cite{NIPS2013_5073} provide no method for estimation and view the noise rates as parameters optimized with cross-validation, inducing a sacrificial accuracy, efficiency trade-off. In comparison, Rank Pruning noise rate estimation is time-efficient, consistent in ideal conditions, and robust to imperfect probability estimation.

\cite{NIPS2013_5073} developed two methods for prediction in the $\tilde{P}\tilde{N}$ setting which modify the loss function. The first method constructs an unbiased estimator of the loss function for the true distribution from the noisy distribution, but the estimator may be non-convex even if the original loss function is convex. If the classifier's loss function cannot be modified directly, this method requires splitting each example in two with class-conditional weights and ensuring split examples are in the same batch during optimization. For these reasons, we instead compare Rank Pruning with their second method (\emph{Nat13}), which constructs a label-dependent loss function such that for 0-1 loss, the minimizers of \emph{Nat13}'s risk and the risk for the true distribution are equivalent. 


\cite{Liu:2016:CNL:2914183.2914328} generalized \emph{Elk08} to the $\tilde{P}\tilde{N}$ learning setting by modifying the loss function with per-example importance reweighting (\emph{Liu16}), but reweighting terms are derived from predicted probabilities which may be sensitive to inexact estimation. To mitigate sensitivity, \cite{Liu:2016:CNL:2914183.2914328} examine the use of density ratio estimation \citep{Sugiyama:2012:DRE:2181148}. Instead, Rank Pruning mitigates sensitivity by learning from confident examples selected by rank order, not predicted probability. For fairness of comparison across methods, we compare Rank Pruning with their probability-based approach.

Assuming perfect estimation of $\rho_1$ and $\rho_0$, we, \cite{NIPS2013_5073}, and \cite{Liu:2016:CNL:2914183.2914328} all prove that the expected risk for the modified loss function is equivalent to the expected risk for the perfectly labeled dataset. However, both \cite{NIPS2013_5073} and \cite{Liu:2016:CNL:2914183.2914328} effectively "flip" example labels in the construction of their loss function, providing no benefit for added random noise. In comparison, Rank Pruning will also remove added random noise because noise drawn from a third distribution is unlikely to appear confidently positive or negative. Table \ref{t:related_works} summarizes our comparison of $\tilde{P}\tilde{N}$ and $PU$ learning methods.


Procedural efforts have improved robustness to mislabeling in the context of machine vision \citep{Xiao2015LearningFM}, neural networks \citep{noisy_boostrapping_google}, and face recognition \citep{angelova2005pruning}. Though promising, these methods are restricted in theoretical justification and generality, motivating the need for Rank Pruning.





\subsection{Contributions}

In this paper, we describe the Rank Pruning algorithm for binary classification with imperfectly labeled training data. In particular, we:

\vskip -0.05in
\begin{itemize}
\vskip -0.05in
\itemsep.05in 
        \item Develop a robust, time-efficient, general solution for both $\tilde{P}\tilde{N}$ learning, i.e. binary classification with noisy labels, and estimation of the fraction of mislabeling in both the positive and negative training sets.
        \item Introduce the \emph{learning with confident examples} mantra as a new way to think about robust classification and estimation with mislabeled training data.
        \item Prove that under assumptions, Rank Pruning achieves perfect noise estimation and equivalent expected risk as learning with correct labels. We provide closed-form solutions when those assumptions are relaxed.
        \item Demonstrate that Rank Pruning performance generalizes across the number of training examples, feature dimension, fraction of mislabeling, and fraction of added noise examples drawn from a third distribution.
        \item Improve the state-of-the-art of $\tilde{P}\tilde{N}$ learning across F1 score, AUC-PR, and Error. In many cases, Rank Pruning achieves nearly the same F1 score as learning with correct labels when 50\% of positive examples are mislabeled and 50\% of observed positive labels are mislabeled negative examples.
    \end{itemize}

\section{Framing the \texorpdfstring{$\tilde{P}\tilde{N}$}{Noisy PN} Learning Problem }
\label{sec:framing}

\vskip -0.04in

In this section, we formalize the foundational definitions, assumptions, and goals of the $\tilde{P}\tilde{N}$ learning problem illustrated by the student-teacher motivational example.

Given $n$ observed training examples $x \in \mathcal{R}^D$ with associated observed corrupted labels $s \in \{0,1\}$ and unobserved true labels $y \in \{0,1\}$, we seek a binary classifier $f$ that estimates the mapping $x\to y$. Unfortunately, if we fit the classifier using observed $(x, s)$ pairs, we estimate the mapping $x\to s$ and obtain $g(x)=P(\hat{s}=1|x)$. 

We define the observed noisy positive and negative sets as $\tilde{P} = \{x| s = 1\}, \tilde{N} = \{x| s = 0\}$ and the unobserved true positive and negative sets as $P = \{x| y = 1\}, N = \{x| y = 0\}$. Define the hidden training data as $D=\{(x_1, y_1), (x_2, y_2), ..., (x_n, y_n)\}$, drawn i.i.d. from some true distribution $\mathcal{D}$. 
We assume that a class-conditional Classification Noise Process (CNP) \citep{angluin1988learning} maps $y$ true labels to $s$ observed labels such that each label in $P$ is flipped independently with probability $\rho_1$ and each label in $N$ is flipped independently with probability $\rho_0$ ($s \leftarrow CNP(y, \rho_1, \rho_0)$). 
The resulting observed, corrupted dataset is $D_{\rho}=\{(x_1, s_1), (x_2, s_2), ..., (x_n, s_n)\}$. Therefore, $(s \independent x) | y$ and $P(s=s|y=y,x) = P(s=s|y=y)$. In recent work, CNP is referred to as the random noise classification (RCN) noise model \citep{Liu:2016:CNL:2914183.2914328, NIPS2013_5073}.

The noise rate $\rho_1 = P(s=0|y=1)$ is the fraction of $P$ examples mislabeled as negative and the noise rate $\rho_0 = P(s=1|y=0)$ is the fraction of $N$ examples mislabeled as positive. Note that $\rho_1 + \rho_0 < 1$ is a necessary condition, otherwise more examples would be mislabeled than labeled correctly. Thus, $\rho_0 < 1- \rho_1$. We elect a subscript of ``0" to refer to the negative set and a subscript of ``1" to refer to the positive set. Additionally, let $p_{s1} = P(s=1)$ be the fraction of corrupted labels that are positive and  $p_{y1} = P(y=1)$ be the fraction of true labels that are positive. It follows that the inversed noise rates are $\pi_1 = P(y=0|s=1) =  \frac{\rho_0 (1 - p_{y1})}{p_{s1}}$ and $\pi_0 = P(y=1|s=0) =  \frac{\rho_1 p_{y1}}{(1 - p_{s1})}$. Combining these relations, given any pair in $\{(\rho_0, \rho_1),(\rho_1, \pi_1),(\rho_0, \pi_0), (\pi_0, \pi_1)\}$, the remaining two and $p_{y1}$ are known. 

We consider five levels of assumptions for $P$, $N$, and $g$: \\
\textbf{Perfect Condition}: $g$ is a ``perfect" probability estimator iff $g(x) = g^*(x)$ where $g^*(x) = P(s=1|x)$. Equivalently, let $g(x) = P(s=1|x)+\Delta g(x)$. Then $g(x)$ is ``perfect" when $\Delta g(x) = 0$ and ``imperfect" when $\Delta g(x) \neq 0$. $g$ may be imperfect due to the method of estimation or due to added uniformly randomly labeled examples drawn from a third noise distribution.\\
\textbf{Non-overlapping Condition}: $P$ and $N$ have ``non-overlapping support" if $P(y=1|x) = \indicator{y=1}$, where the indicator function $\indicator{a}$ is $1$ if the $a$ is true, else $0$. \\
\textbf{Ideal Condition}\footnote{ Eq. (\ref{eq1}) is first derived in \citep{Elkan:2008:LCO:1401890.1401920} .}: $g$ is ``ideal" when both perfect and non-overlapping conditions hold and $(s \independent x) | y$ such that

\vskip -0.25in
\begin{equation} \label{eq1}
\begin{split}
g(x) = & g^*(x) = P(s=1|x) \\ 
 = & P(s=1|y=1,x) \cdot P(y=1|x) + P(s=1|y=0,x) \cdot P(y=0|x) \\
 = & (1 - \rho_1) \cdot \mathbbm{1}[[y=1]] + \rho_0 \cdot \mathbbm{1}[[y=0]]
\end{split}
\end{equation}
\vskip -0.1in

\textbf{Range Separability Condition} $g$ range separates $P$ and $N$ iff $\forall x_1\in P$ and $\forall x_2 \in N$, we have $g(x_1) >g(x_2)$. \\
\textbf{Unassuming Condition}: $g$ is ``unassuming" when perfect and/or non-overlapping conditions may not be true. 

Their relationship is:
$\textbf{Unassuming}\supset\textbf{Range}$ $\textbf{Separability}$ $\supset\textbf{Ideal}=\textbf{Perfect}\cap\textbf{Non-overlapping}$. 

We can now state the two goals of Rank Pruning for $\tilde{P}\tilde{N}$ learning. \textbf{Goal 1} is to perfectly estimate $\hat{\rho}_1 \estimates \rho_1$ and $\hat{\rho}_0  \estimates \rho_0$ when $g$ is ideal. When $g$ is not ideal, to our knowledge perfect estimation of $\rho_1$ and $\rho_0$ is impossible and at best \textbf{Goal 1} is to provide exact expressions for $\hat{\rho}_1$ and $\hat{\rho}_0$ w.r.t. $\rho_1$ and $\rho_0$. \textbf{Goal 2} is to use $\hat{\rho}_1$ and $\hat{\rho}_0$ to uncover the classifications of $f$ from $g$. Both tasks must be accomplished given only observed ($x,s$) pairs. $y, \rho_1, \rho_0, \pi_1$, and $\pi_0$ are hidden.


\section{Rank Pruning}\label{methodology}

We develop the Rank Pruning algorithm to address our two goals. In Section \ref{method:noise}, we propose a method for noise rate estimation and prove consistency when $g$ is ideal. An estimator is ``consistent" if it achieves perfect estimation in the expectation of infinite examples. In Section \ref{method:noise_non_ideal}, we derive exact expressions for $\hat{\rho}_1$ and $\hat{\rho}_0$ when $g$ is unassuming. In Section \ref{method:rp}, we provide the entire algorithm, and in Section \ref{method:risk}, prove that Rank Pruning has equivalent expected risk as learning with uncorrupted labels for both ideal $g$ and non-ideal $g$ with weaker assumptions. Throughout, we assume $n \rightarrow \infty$ so that $P$ and $N$ are the hidden distributions, each with infinite examples. This is a necessary condition for Theorems. \ref{thm:noise_ideal}, \ref{rho_conf_robust} and Lemmas \ref{thm:lemma1}, \ref{lemma3}.

\subsection{Deriving Noise Rate Estimators \texorpdfstring{$\hat{\rho}_1^{conf}$ and $\hat{\rho}_0^{conf}$}{}} \label{method:noise}

We propose the \emph{confident counts} estimators $\hat{\rho}_1^{conf}$ and $\hat{\rho}_0^{conf}$ to estimate $\rho_1$ and $\rho_0$ as a fraction of the predicted counts of confident examples in each class, encouraging robustness for variation in probability estimation. To estimate $\rho_1=P(s=0|y=1)$, we count the number of examples with label $s=0$ that we are ``confident" have label $y=1$ and divide it by the total number of examples that we are ``confident" have label $y = 1$. More formally, 

\vskip -0.2in
\begin{equation}\label{define_rho_conf}
\hat{\rho}_1^{conf}:=\frac{|\tilde{N}_{y=1}|}{|\tilde{N}_{y=1}|+|\tilde{P}_{y=1}|}, \qquad
\hat{\rho}_0^{conf}:=\frac{|\tilde{P}_{y=0}|}{|\tilde{P}_{y=0}|+|\tilde{N}_{y=0}|}
\end{equation}
\vskip -0.1in

such that

\vskip -0.05in
\begin{equation}\label{define_threshold}
\begin{cases}
\tilde{P}_{y=1}=\{x\in\tilde{P}\mid g(x)\geq LB_{y=1}\}\\
\tilde{N}_{y=1}=\{x\in\tilde{N}\mid g(x)\geq LB_{y=1}\}\\
\tilde{P}_{y=0}=\{x\in\tilde{P}\mid g(x)\leq UB_{y=0}\}\\
\tilde{N}_{y=0}=\{x\in\tilde{N}\mid g(x)\leq UB_{y=0}\}
\end{cases}
\end{equation}
\vskip -0.1in

where $g$ is fit to the corrupted training set $D_{\rho}$ to obtain $g(x)=P(\hat{s}=1|x)$. The threshold $LB_{y=1}$ is the predicted probability in $g(x)$ above which we guess that an example $x$ has hidden label $y = 1$, and similarly for upper bound $UB_{y=0}$. $LB_{y=1}$ and $UB_{y=0}$ partition $\tilde{P}$ and $\tilde{N}$ into four sets representing a \emph{best guess} of a \emph{subset} of examples having labels (1) $s=1, y=0$, (2) $s=1, y=1$, (3) $s=0, y=0$, (4) $s=0, y=1$. The threshold values are defined as

\vskip -0.15in
\begin{equation*}
\begin{cases}
LB_{y=1}:=P(\hat{s}=1\mid s=1)=E_{x\in\tilde{P}}[g(x)]\\
UB_{y=0}:=P(\hat{s}=1\mid s=0)=E_{x\in\tilde{N}}[g(x)]\\
\end{cases}
\end{equation*}
\vskip -0.1in

where $\hat{s}$ is the predicted label from a classifier fit to the observed data. $|\tilde{P}_{y=1}|$ counts examples with label $s=1$ that are \emph{most} likely to be correctly labeled ($y=1$) because $LB_{y=1} = P(\hat{s}=1|s=1)$. The three other terms in Eq. (\ref{define_threshold}) follow similar reasoning. Importantly, the four terms do not sum to $n$, i.e. $|N| + |P|$, but $\hat{\rho}_1^{conf}$ and $\hat{\rho}_0^{conf}$ are valid estimates because mislabeling noise is assumed to be uniformly random. 
The choice of threshold values relies on the following two important equations:

\vskip -0.25in
\begin{align*}
LB_{y=1}=&E_{x\in\tilde{P}}[g(x)]=E_{x\in\tilde{P}}[P(s=1|x)]\\
=&E_{x\in\tilde{P}}[P(s=1|x,y=1)P(y=1|x)+P(s=1|x,y=0)P(y=0|x)]\\
=&E_{x\in\tilde{P}}[P(s=1|y=1)P(y=1|x)+P(s=1|y=0)P(y=0|x)]\\
=&(1-\rho_1)(1-\pi_1)+\rho_0\pi_1\numberthis \label{rh1_prob_eq}
\end{align*}
\vskip -0.05in

Similarly, we have

\vskip -0.15in
\begin{equation}\label{rh0_prob_eq}
UB_{y=0}=(1-\rho_1)\pi_0+\rho_0(1-\pi_0)
\end{equation}
\vskip -0.05in

To our knowledge, although simple, this is the first time that the relationship in Eq. (\ref{rh1_prob_eq}) (\ref{rh0_prob_eq}) has been published, linking the work of \cite{Elkan:2008:LCO:1401890.1401920}, \cite{Liu:2016:CNL:2914183.2914328}, \cite{ScottBH13} and \cite{NIPS2013_5073}. From Eq. (\ref{rh1_prob_eq}) (\ref{rh0_prob_eq}), we observe that $LB_{y=1}$ and $UB_{y=0}$ are linear interpolations of $1-\rho_1$ and $\rho_0$ and since $\rho_0 < 1- \rho_1$, we have that $\rho_0 < LB_{y=1} \leq 1-\rho_1$ and $\rho_0 \leq UB_{y=0} < 1-\rho_1$. When $g$ is ideal we have that $g(x)=(1-\rho_1)$, if $x \in P$ and $g(x)=\rho_0$, if $x \in N$. Thus  when $g$ is ideal, the thresholds $LB_{y=1}$ and  $UB_{y=0}$ in Eq. (\ref{define_threshold}) will perfectly separate $P$ and $N$ examples within each of $\tilde{P}$ and $\tilde{N}$. Lemma \ref{thm:lemma1} immediately follows.

\begin{lemma}\label{thm:lemma1}
\vskip -.1in
When $g$ is ideal,
\vskip -0.25in

\begin{align*}
\tilde{P}_{y=1} &= \{x\in P\mid s=1\},
\tilde{N}_{y=1} = \{x\in P\mid s=0\},\\
\tilde{P}_{y=0} &= \{x\in N\mid s=1\},
\tilde{N}_{y=0} = \{x\in N\mid s=0\}\numberthis
\end{align*}
\vskip -0.3in
\end{lemma}
\vskip -0.15in

Thus, when $g$ is ideal, the thresholds in Eq. (\ref{define_threshold}) partition the training set such that $\tilde{P}_{y=1}$ and $\tilde{N}_{y=0}$ contain the correctly labeled examples and $\tilde{P}_{y=0}$ and $\tilde{N}_{y=1}$ contain the mislabeled examples. Theorem \ref{thm:noise_ideal} follows (for brevity, proofs of all theorems/lemmas are in Appendix \ref{sec:A1}-\ref{sec:A5}).

\begin{theorem} \label{thm:noise_ideal}

When $g$ is ideal,
\vskip -0.15in
\begin{equation}
\hat{\rho}_1^{conf}=\rho_1, \hat{\rho}_0^{conf}=\rho_0
\end{equation}
\end{theorem}

Thus, when $g$ is ideal, the \emph{confident counts} estimators $\hat{\rho}_1^{conf}$ and $\hat{\rho}_0^{conf}$ are consistent estimators for $\rho_1$ and $\rho_0$ and we set $\hat{\rho}_1:=\hat{\rho}_1^{conf}, \hat{\rho}_0:=\hat{\rho}_0^{conf}$. These steps comprise Rank Pruning noise rate estimation (see Alg. \ref{alg:rp}). There are two practical observations. First, for any $g$ with $T$ fitting time, computing $\hat{\rho}_1^{conf}$ and $\hat{\rho}_0^{conf}$ is $\mathcal{O}(T)$. Second, $\hat{\rho}_1$ and $\hat{\rho}_0$ should be estimated out-of-sample to avoid over-fitting, resulting in sample variations. In our experiments, we use 3-fold cross-validation, requiring at most $2T=\mathcal{O}(T)$. 

\subsection{Noise Estimation: Unassuming Case} \label{method:noise_non_ideal}
 
Theorem \ref{thm:noise_ideal} states that $\hat{\rho}_i^{conf} = \rho_i$, $\forall i \in \{0,1\}$ when $g$ is ideal. Though theoretically constructive, in practice this is unlikely. Next, we derive expressions for the estimators when $g$ is unassuming, i.e. $g$ may not be perfect and $P$ and $N$ may have overlapping support.

Define $\Delta p_o:=\frac{|P \cap N|}{|P \cup N|}$ as the fraction of overlapping examples in $\mathcal{D}$ and remember that $\Delta g(x) := g(x)-g^*(x)$. Denote $ LB_{y=1}^*=(1-\rho_1)(1-\pi_1)+\rho_0\pi_1, UB_{y=0}^*=(1-\rho_1)\pi_0+\rho_0(1-\pi_0)$. We have

\begin{lemma}\label{lemma3}
When $g$ is unassuming, we have

\vskip -0.1in
\begin{equation}\label{rho_imperfect_condition}
\begin{cases}
LB_{y=1}=LB_{y=1}^*+E_{x\in\tilde{P}}[\Delta g(x)]-\frac{(1-\rho_1-\rho_0)^2}{p_{s1}}\Delta p_o\\
UB_{y=0}=UB_{y=0}^*+E_{x\in\tilde{N}}[\Delta g(x)]+\frac{(1-\rho_1-\rho_0)^2}{1-p_{s1}}\Delta p_o\\
\hat{\rho}_1^{conf}=\rho_1+\frac{1-\rho_1-\rho_0}{|P|-|\Delta P_1| + |\Delta N_1|}|\Delta N_1|\\
\hat{\rho}_0^{conf}=\rho_0+\frac{1-\rho_1-\rho_0}{|N|-|\Delta N_0| + |\Delta P_0|}|\Delta P_0|
\end{cases}
\end{equation}
\vskip -0.1in

where

\vskip -0.15in
\begin{equation*}
\begin{cases}
\Delta P_1=\{x\in P\mid g(x)< LB_{y=1}\}\\
\Delta N_1=\{x\in N\mid g(x)\geq LB_{y=1}\}\\
\Delta P_0=\{x\in P\mid g(x)\leq UB_{y=0}\}\\
\Delta N_0=\{x\in N\mid g(x)> UB_{y=0}\}\\
\end{cases}
\end{equation*}
\vskip -0.35in
\end{lemma}
\vskip -0.1in

The second term on the R.H.S. of the $\hat{\rho}_i^{conf}$ expressions captures the deviation of $\hat{\rho}_i^{conf}$ from $\rho_i$, $i=0,1$. This term results from both imperfect $g(x)$ and overlapping support. Because the term is non-negative, $\hat{\rho}_i^{conf} \geq \rho_i$, $i=0,1$ in the limit of infinite examples. In other words, $\hat{\rho}_i^{conf}$ is an \emph{upper bound} for the noise rates $\rho_i$, $i=0,1$. From Lemma \ref{lemma3}, it also follows:

\begin{theorem}\label{rho_conf_robust}
Given non-overlapping support condition,

\text{If} $\forall x\in N, \Delta g(x)<LB_{y=1}-\rho_0$, then $\hat{\rho}_1^{conf}=\rho_1$.

If $\forall x\in P, \Delta g(x)>-(1-\rho_1-UB_{y=0}$), then $\hat{\rho}_0^{conf}=\rho_0$.
\end{theorem}

Theorem \ref{rho_conf_robust} shows that $\hat{\rho}_1^{conf}$ and $\hat{\rho}_0^{conf}$ are robust to imperfect probability estimation. As long as $\Delta g(x)$ does not exceed the distance between the threshold in Eq. (\ref{define_threshold}) and the perfect $g^*(x)$ value, $\hat{\rho}_1^{conf}$ and $\hat{\rho}_0^{conf}$ are consistent estimators for $\rho_1$ and $\rho_0$. Our numerical experiments in Section \ref{sec:experimental} suggest this is reasonable for $\Delta g(x)$. The average $|\Delta g(x)|$ for the MNIST training dataset across different ($\rho_1$, $\pi_1$) varies between 0.01 and 0.08 for a logistic regression classifier, 0.01$\sim$0.03 for a CNN classifier, and 0.05$\sim$0.10 for the CIFAR dataset with a CNN classifier. Thus, when  $LB_{y=1}-\rho_0$ and $1-\rho_1-UB_{y=0}$ are above 0.1 for these datasets, from Theorem  \ref{rho_conf_robust} we see that $\hat{\rho}_i^{conf}$ still accurately estimates $\rho_i$.

\subsection{The Rank Pruning Algorithm} \label{method:rp}

Using $\hat{\rho}_1$ and $\hat{\rho}_0$, we must uncover the classifications of $f$ from $g$. In this section, we describe how Rank Pruning selects confident examples, removes the rest, and trains on the pruned set using a reweighted loss function. 

First, we obtain the inverse noise rates $\hat{\pi}_1$, $\hat{\pi}_0$ from $\hat{\rho}_1$, $\hat{\rho}_0$:

\vskip -0.13in
\begin{equation}
\hat{\pi}_1=\frac{\hat{\rho}_0}{p_{s1}}\frac{1-p_{s1}-\hat{\rho}_1}{1-\hat{\rho}_1-\hat{\rho}_0},\qquad
\hat{\pi}_0=\frac{\hat{\rho}_1}{1-p_{s1}}\frac{p_{s1}-\hat{\rho}_0}{1-\hat{\rho}_1-\hat{\rho}_0}
\end{equation}
\vskip -0.07in

Next, we prune the $\hat{\pi}_1|\tilde{P}|$ examples in $\tilde{P}$ with smallest $g(x)$ and the $\hat{\pi}_0|\tilde{N}|$ examples in $\tilde{N}$ with highest $g(x)$ and denote the pruned sets $\tilde{P}_{conf}$ and $\tilde{N}_{conf}$. To prune, we define $k_1$ as the $(\hat{\pi}_1|\tilde{P}|)^{th}$ smallest $g(x)$ for $x \in \tilde{P}$ and $k_0$ as the $(\hat{\pi}_0|\tilde{N}|)^{th}$ largest $g(x)$ for $x \in \tilde{N}$. BFPRT ($\mathcal{O}(n)$) \citep{Blum:1973:TBS:1739940.1740109} is used to compute $k_1$ and $k_0$ and pruning is reduced to the following $\mathcal{O}(n)$ filter:

\vskip -0.13in
\begin{equation}
\tilde{P}_{conf} := \{x \in \tilde{P} \mid g(x) \geq k_1 \},\qquad 
\tilde{N}_{conf} := \{x \in \tilde{N} \mid g(x) \leq k_0 \}
\end{equation}
\vskip -0.05in

\begin{figure}[t]
\begin{center}
\centerline{\includegraphics[width=1.0\columnwidth]{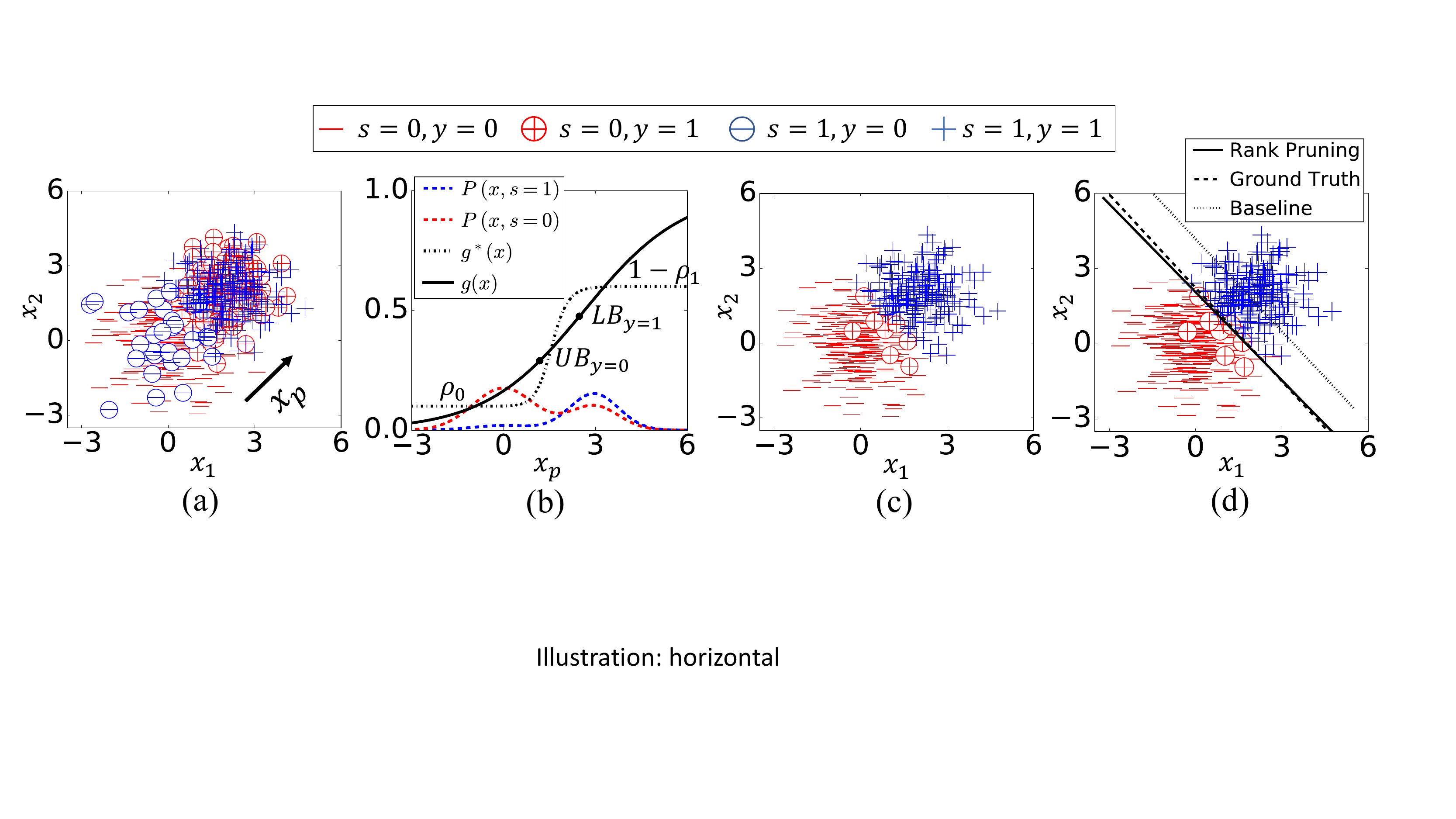}}
\caption{Illustration of Rank Pruning with a logistic regression classifier ($\mathcal{LR}_\theta$).  \textbf{(a)}: The corrupted training set $D_\rho$ with noise rates $\rho_1=0.4$ and $\rho_0=0.1$. Corrupted colored labels (\textcolor{blue}{$s=1$},\textcolor{red}{$s=0$}) are observed. $y$ ($+$,$-$) is hidden. \textbf{(b)}: The marginal distribution of $D_\rho$ projected onto the $x_p$ axis (indicated in (a)), and the $\mathcal{LR}_\theta$'s estimated $g(x)$, from which $\hat{\rho}_1^{conf}=0.4237$, $\hat{\rho}_0^{conf}=0.1144$ are estimated. \textbf{(c)}: The pruned $X_{conf}, s_{conf}$. \textbf{(d)}: The classification result by Rank Pruning ($\hat{f}$ = $\mathcal{LR}_\theta.\text{fit}(X_{conf}, s_{conf})$),
ground truth classifier ($f$ = $\mathcal{LR}_\theta.\text{fit}(X,y)$),
and baseline classifier ($g$ = $\mathcal{LR}_\theta.\text{fit}(X, s)$), with an accuracy of $94.16\%$, $94.16\%$ and $78.83\%$, respectively.}
\label{rankpruning_illustration}
\end{center}
\vskip -0.15in
\end{figure}

Lastly, we refit the classifier to $X_{conf} = \tilde{P}_{conf} \cup \tilde{N}_{conf}$ by class-conditionally reweighting the loss function for examples in $\tilde{P}_{conf}$ with weight $\frac{1}{1-\hat{\rho}_1}$ and examples in $\tilde{N}_{conf}$ with weight $\frac{1}{1-\hat{\rho}_0}$ to recover the estimated balance of positive and negative examples. The entire Rank Pruning algorithm is presented in Alg. \ref{alg:rp} and illustrated step-by-step on a synthetic dataset in Fig. \ref{rankpruning_illustration}.

We conclude this section with a formal discussion of the loss function and efficiency of Rank Pruning.  Define $\hat{y_i}$ as the predicted label of example $i$ for the classifier fit to $X_{conf}, s_{conf}$ and let $l(\hat{y_i}, s_i)$ be the original loss function for $x_i \in D_\rho$. Then the loss function for Rank Pruning is simply the original loss function exerted on the pruned $X_{conf}$, with class-conditional weighting:

\vskip -0.25in
\begin{align*}  \label{rank_pruning_loss_function}
\tilde{l}(\hat{y_i}, s_i)=&\frac{1}{1-\hat{\rho}_1}l(\hat{y_i}, s_i)\cdot\indicator{x_i\in \tilde{P}_{conf}}+ \frac{1}{1-\hat{\rho}_0}l(\hat{y_i}, s_i)\cdot\indicator{x_i\in \tilde{N}_{conf}}\numberthis 
\end{align*}

\medskip

Effectively this loss function uses a zero-weight for pruned examples. Other than potentially fewer examples, the only difference in the loss function for Rank Pruning and the original loss function is the class-conditional weights. These constant factors do not increase the complexity of the minimization of the original loss function. In other words, we can fairly report the running time of Rank Pruning in terms of the running time ($\mathcal{O}(T)$) of the choice of probabilistic estimator. Combining noise estimation ($\mathcal{O}(T)$), pruning ($\mathcal{O}(n)$), and the final fitting ($\mathcal{O}(T)$), Rank Pruning has a running time of $\mathcal{O}(T) + \mathcal{O}(n)$, which is $\mathcal{O}(T)$ for typical classifiers.

\subsection{Rank Pruning: A simple summary}

Recognizing that formalization can create obfuscation, in this section we describe the entire algorithm in a few sentences. Rank Pruning takes as input training examples $X$, noisy labels $s$, and a probabilistic classifier $clf$ and finds a subset of $X, s$ that is likely to be correctly labeled, i.e. a subset of $X, y$. To do this, we first find two thresholds, $LB_{y=1}$ and $UB_{y=0}$, to \emph{confidently} guess the correctly and incorrectly labeled examples in each of $\tilde{P}$ and $\tilde{N}$, forming four sets, then use the set sizes to estimate the noise rates $\rho_1 = P(s = 0 | y = 1)$ and $\rho_0 = P(s = 1 | y = 0)$. We then use the noise rates to estimate the number of examples with observed label $s = 1$ and hidden label $y = 0$ and remove that number of examples from $\tilde{P}$ by removing those with lowest predicted probability $g(x)$. We prune $\tilde{N}$ similarly. Finally, the classifier is fit to the pruned set, which is intended to represent a subset of the correctly labeled data.

\begin{algorithm}[t]
   \caption{\textbf{Rank Pruning}}
   \label{alg:rp}
\begin{algorithmic}
   \STATE {\bfseries Input:} Examples $X$, corrupted labels $s$, classifier clf
   \STATE \textbf{Part 1. Estimating Noise Rates:}
   \STATE (1.1)\ \ clf.fit($X$,$s$)
   \STATE \ \ \ \ \ \ \ \ \ $g(x)\leftarrow$clf.predict\_crossval\_probability($\hat{s}=1|x$)
   \STATE \ \ \ \ \ \  \ \ \ $p_{s1}=\frac{\text{count}(s=1)}{\text{count}(s=0 \lor s=1)}$
   \STATE \ \ \ \ \ \  \ \ \ $LB_{y=1}=E_{x\in\tilde{P}}[g(x)]$, $UB_{y=0}=E_{x\in\tilde{N}}[g(x)]$
   \STATE (1.2)\  $\hat{\rho}_1=\hat{\rho}_1^{conf}=\frac{|\tilde{N}_{y=1}|}{|\tilde{N}_{y=1}|+|\tilde{P}_{y=1}|}$, $\hat{\rho}_0=\hat{\rho}_0^{conf}=\frac{|\tilde{P}_{y=0}|}{|\tilde{P}_{y=0}|+|\tilde{N}_{y=0}|}$
   \STATE \ \ \ \ \ \ \ \ \ $\hat{\pi}_1=\frac{\hat{\rho}_0}{p_{s1}}\frac{1-p_{s1}-\hat{\rho}_1}{1-\hat{\rho}_1-\hat{\rho}_0}$, $\hat{\pi}_0=\frac{\hat{\rho}_1}{1-p_{s1}}\frac{p_{s1}-\hat{\rho}_0}{1-\hat{\rho}_1-\hat{\rho}_0}$
   \vskip .1in
   \STATE \textbf{Part 2. Prune Inconsistent Examples:}
   \STATE (2.1) Remove $\hat{\pi}_1|\tilde{P}|$ examples in $\tilde{P}$ with least $g(x)$, Remove $\hat{\pi}_0|\tilde{N}|$ examples in $\tilde{N}$ with greatest $g(x)$,
   \STATE \ \ \ \ \ \ \ \ \ Denote the remaining training set ($X_{conf}$, $s_{conf}$)
   \STATE (2.2) clf.fit($X_{conf}$, $s_{conf}$), with sample weight 
   $w(x)=\frac{1}{1-\hat{\rho}_1}\indicator{s_{conf}=1}$+$\frac{1}{1-\hat{\rho}_0}\indicator{s_{conf}=0}$
   \STATE {\bfseries Output:} clf
\end{algorithmic}
\end{algorithm}

\subsection{Expected Risk Evaluation} 
\label{method:risk}

In this section, we prove Rank Pruning exactly uncovers the classifier $f$ fit to hidden $y$ labels when $g$ range separates $P$ and $N$ and $\rho_1$ and $\rho_0$ are given.

Denote $f_{\theta}\in\mathcal{F}: x\to \hat{y}$ as a classifier's prediction function belonging to some function space $\mathcal{F}$, where $\theta$ represents the classifier's parameters. $f_{\theta}$ represents $f$, but without $\theta$ necessarily fit to the training data. $\hat{f}$ is the Rank Pruning estimate of $f$.


Denote the empirical risk of $f_{\theta}$ w.r.t. the loss function $\tilde{l}$ and corrupted data $D_{\rho}$ as $\hat{R}_{\tilde{l}, D_{\rho}}(f_{\theta})=\frac{1}{n}\sum_{i=1}^n\tilde{l}(f_{\theta}(x_i), s_i)$, and the expected risk of $f_{\theta}$ w.r.t. the corrupted distribution $\mathcal{D}_{\rho}$ as $R_{\tilde{l},\mathcal{D}_{\rho}}(f_{\theta})=E_{(x,s)\sim \mathcal{D}_{\rho}}[\hat{R}_{\tilde{l},\mathcal{D}_{\rho}}(f_{\theta})]$. Similarly, denote $R_{l,\mathcal{D}}(f_{\theta})$ as the expected risk of $f_{\theta}$ w.r.t. the hidden distribution $\mathcal{D}$ and loss function $l$. We show that using Rank Pruning, a classifier $\hat{f}$ can be learned for the hidden data $D$, given the corrupted data $D_{\rho}$, by minimizing the empirical risk:

\vskip -0.12in
\begin{equation}\label{risk_minimization}
\hat{f}=\argmin\limits_{f_{\theta}\in \mathcal{F}} \hat{R}_{\tilde{l},D_{\rho}}(f_{\theta})=\argmin\limits_{f_{\theta}\in \mathcal{F}} \frac{1}{n}\sum_{i=1}^n\tilde{l}(f_{\theta}(x_i), s_i)
\end{equation}

Under the \emph{range separability} condition, we have

\begin{theorem}\label{Rank Pruning_with_range_separability}
If $g$ range separates $P$ and $N$ and $\hat{\rho}_i=\rho_i$, $i=0,1$, then for any classifier $f_{\theta}$ and any bounded loss function $l(\hat{y}_i,y_i)$, we have

\vskip -0.12in
\begin{equation}
R_{\tilde{l},\mathcal{D}_{\rho}}(f_{\theta})=R_{l,\mathcal{D}}(f_{\theta})
\end{equation}
\vskip -0.06in

where $\tilde{l}(\hat{y}_i,s_i)$ is Rank Pruning's loss function (Eq. \ref{rank_pruning_loss_function}).
\end{theorem}

The proof of Theorem \ref{Rank Pruning_with_range_separability} is in Appendix \ref{sec:A5}. Intuitively, Theorem \ref{Rank Pruning_with_range_separability} tells us that if $g$ range separates $P$ and $N$, then given exact noise rate estimates, Rank Pruning will exactly prune out the positive examples in $\tilde{N}$ and negative examples in $\tilde{P}$, leading to the same expected risk as learning from uncorrupted labels. Thus, Rank Pruning can exactly uncover the classifications of $f$ (with infinite examples) because the expected risk is equivalent for any $f_{\theta}$. Note Theorem \ref{Rank Pruning_with_range_separability} also holds when $g$ is ideal, since \emph{ideal} $\subset$ \emph{range separability}. In practice, \emph{range separability} encompasses a wide range of imperfect $g(x)$ scenarios, e.g. $g(x)$ can have large fluctuation in both $P$ and $N$ or have systematic drift w.r.t. to $g^*(x)$ due to underfitting. 


\section{Experimental Results}
\label{sec:experimental}

In Section \ref{methodology}, we developed a theoretical framework for Rank Pruning, proved exact noise estimation and equivalent expected risk when conditions are ideal, and derived closed-form solutions when conditions are non-ideal. Our theory suggests that, in practice, Rank Pruning should (1) accurately estimate $\rho_1$ and $\rho_0$, (2) typically achieve as good or better F1, error and AUC-PR \citep{Davis:2006:RPR:1143844.1143874} as state-of-the-art methods, and (3) be robust to both mislabeling and added noise. 

In this section, we support these claims with an evaluation of the comparative performance of Rank Pruning in non-ideal conditions across thousands of scenarios. These include less complex (MNIST) and more complex (CIFAR) datasets,  simple (logistic regression) and complex (CNN) classifiers, the range of noise rates, added random noise, separability of $P$ and $N$, input dimension, and number of training examples to ensure that Rank Pruning is a general, agnostic solution for $\tilde{P}\tilde{N}$ learning.

\begin{figure}[t]
\begin{center}
\centerline{\includegraphics[width=1.0\columnwidth]{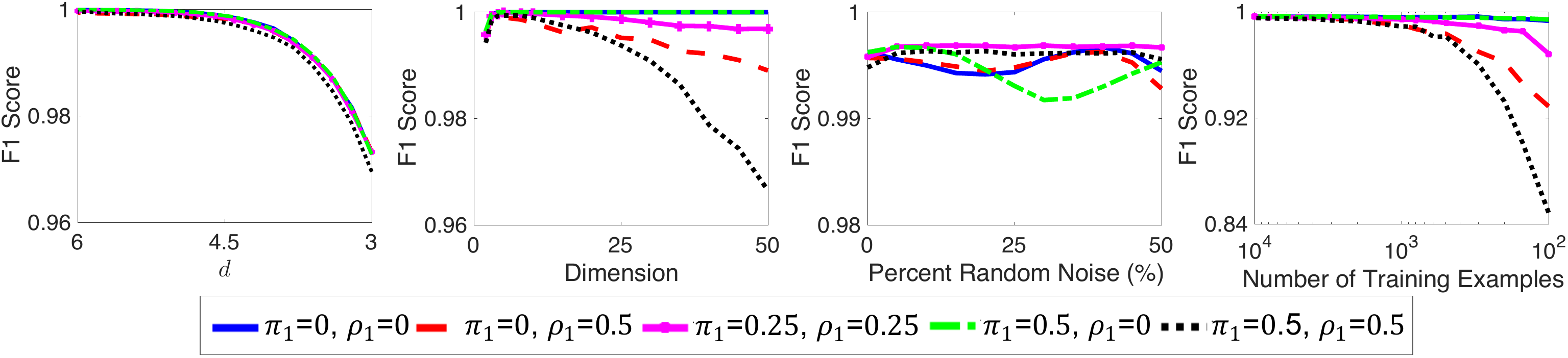}}
\caption{Comparison of Rank Pruning with different noise ratios $(\pi_1, \rho_1)$ on a synthetic dataset for varying separability $d$, dimension, added random noise and number of training examples. Default settings for Fig. \ref{synthetic_F1}, \ref{synthetic_diff_tp} and \ref{synthetic_comparison}: $d=4$, 2-dimension, $0\%$ random noise, and 5000 training examples with $p_{y1}=0.2$. The lines are an average of 200 trials.}
\label{synthetic_F1}
\end{center}
\vskip -0.3in
\end{figure}

\begin{figure}[t]
\begin{center}
\centerline{\includegraphics[width=0.75\columnwidth]{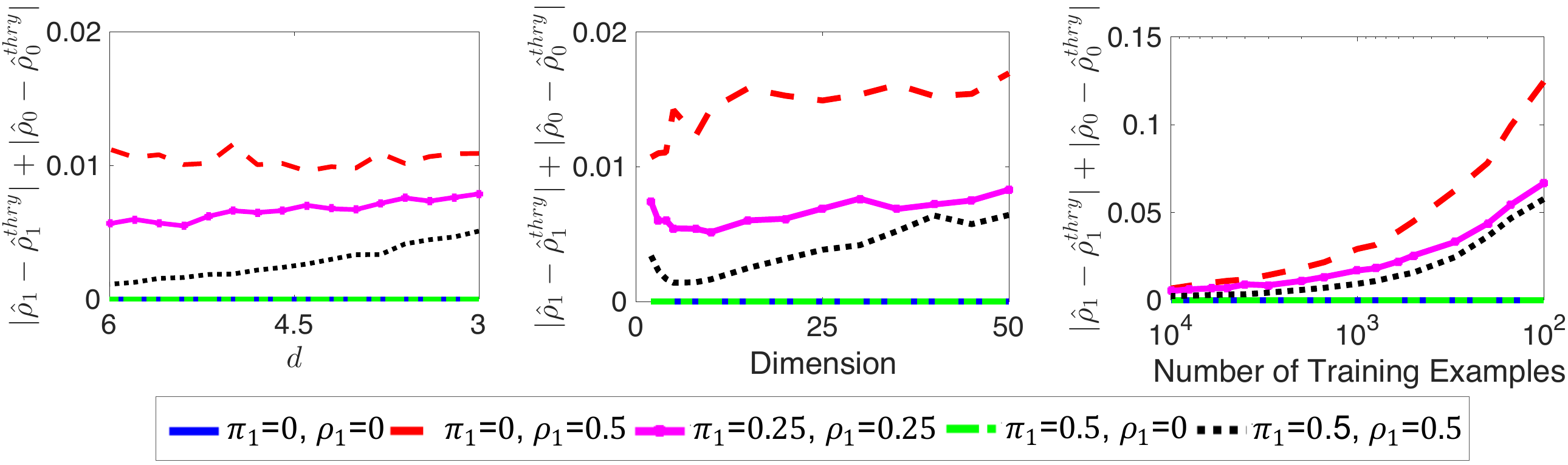}}
\caption{Sum of absolute difference between theoretically estimated $\hat{\rho}_i^{thry}$ and empirical $\hat{\rho}_i$, $i=0,1$, with five different $(\pi_1, \rho_1)$, for varying separability $d$, dimension, and number of training examples. Note that no figure exists for percent random noise because the theoretical estimates in Eq. (\ref{rho_imperfect_condition}) do not address added noise examples.}
\label{synthetic_diff_tp}
\end{center}
\vskip -0.3in
\end{figure}

In our experiments, we adjust $\pi_1$ instead of $\rho_0$ because binary noisy classification problems (e.g. detection and recognition tasks) often have that $|P| \ll |N|$. This choice allows us to adjust both noise rates with respect to $P$, i.e. the fraction of true positive examples that are mislabeled as negative ($\rho_1$) and the fraction of observed positive labels that are actually mislabeled negative examples ($\pi_1$). The $\tilde{P}\tilde{N}$ learning algorithms are trained with corrupted labels $s$, and tested on an unseen test set by comparing predictions $\hat{y}$ with the true test labels $y$ using F1 score, error, and AUC-PR metrics. We include all three to emphasize our apathy toward tuning results to any single metric. We provide F1 scores in this section with error and AUC-PR scores in Appendix \ref{sec:tables_appendix}.

\subsection{Synthetic Dataset}
The synthetic dataset is comprised of a Guassian positive class and a Guassian negative classes such that negative examples ($y=0$) obey an $m$-dimensional Gaussian distribution $N(\mathbf{0}, \mathbf{I})$ with unit variance $\mathbf{I}=diag(1,1,...1)$, and positive examples obey $N(d\mathbf{1}, 0.8\mathbf{I})$, where $d\mathbf{1}=(d,d,...d)$ is an $m$-dimensional vector, and $d$ measures the separability of the positive and negative set.

We test Rank Pruning by varying 4 different settings of the environment: separability $d$, dimension, number of training examples $n$, and percent (of $n$) added random noise drawn from a uniform distribution $U([-10,10]^m)$. In each scenario, we test 5 different $(\pi_1,\rho_1)$ pairs: $(\pi_1,\rho_1)\in\{(0,0),(0,0.5),(0.25,0.25)$, $(0.5,0),(0.5,0.5)\}$. From Fig. \ref{synthetic_F1}, we observe that across these settings, the F1 score for Rank Pruning is fairly agnostic to magnitude of mislabeling (noise rates). As a validation step, in Fig. \ref{synthetic_diff_tp} we measure how closely our empirical estimates match our theoretical solutions in Eq. (\ref{rho_imperfect_condition}) and find near equivalence except when the number of training examples approaches zero.

\begin{figure}[t]
\begin{center}
\centerline{\includegraphics[width=1.0\columnwidth]{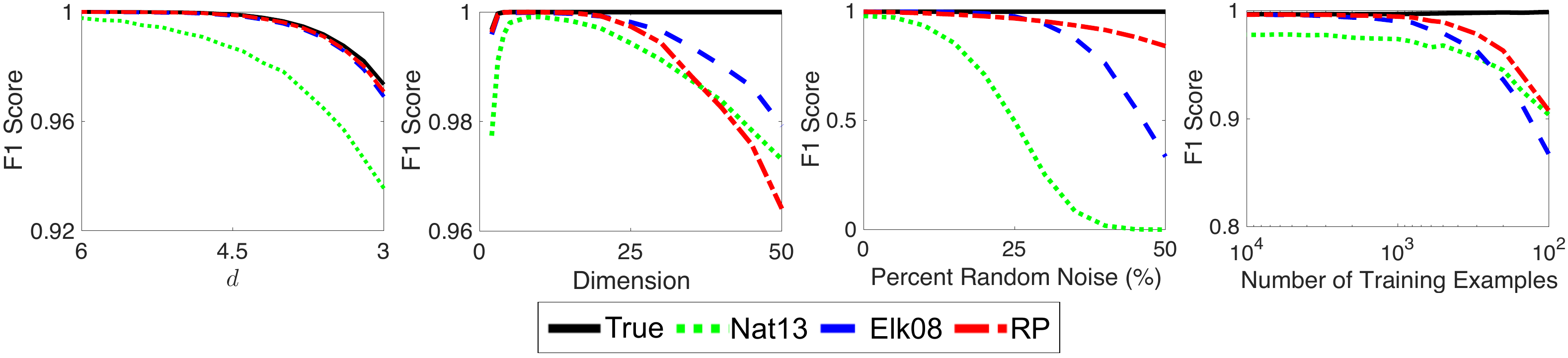}}
\caption{Comparison of $\tilde{P}\tilde{N}$ methods for varying separability $d$, dimension, added random noise, and number of training examples for $\pi_1=0.5$, $\rho_1=0.5$ (given to all methods).
}
\label{synthetic_comparison}
\end{center}
\vskip -0.15in
\end{figure}

For significant mislabeling ($\rho_1=0.5$, $\pi_1=0.5$), Rank Pruning often outperforms other methods (Fig. \ref{synthetic_comparison}). In the scenario of different separability $d$, it achieves nearly the same F1 score as the ground truth classifier. Remarkably, from Fig. \ref{synthetic_F1} and Fig. \ref{synthetic_comparison}, we observe that when added random noise comprises $50\%$ of total training examples, Rank Pruning still achieves F1 $>$ 0.85, compared with F1 $<$ 0.5 for all other methods. This emphasizes a unique feature of Rank Pruning, it will also remove added random noise because noise drawn from a third distribution is unlikely to appear confidently positive or negative.

\subsection{MNIST and CIFAR Datasets}

We consider the binary classification tasks of one-vs-rest for the MNIST \citep{lecun-mnisthandwrittendigit-2010} and CIFAR-10 (\cite{cifar10}) datasets, e.g. the ``car vs rest" task in CIFAR is to predict if an image is a ``car" or ``not". $\rho_1$ and $\pi_1$ are given to all $\tilde{P}\tilde{N}$ learning methods for fair comparison, except for $RP_\rho$ which is Rank Pruning including noise rate estimation. $RP_\rho$ metrics measure our performance on the unadulterated $\tilde{P}\tilde{N}$ learning problem.

\begin{figure}[b!]
\begin{center}
\centerline{\includegraphics[width=1.0\columnwidth]{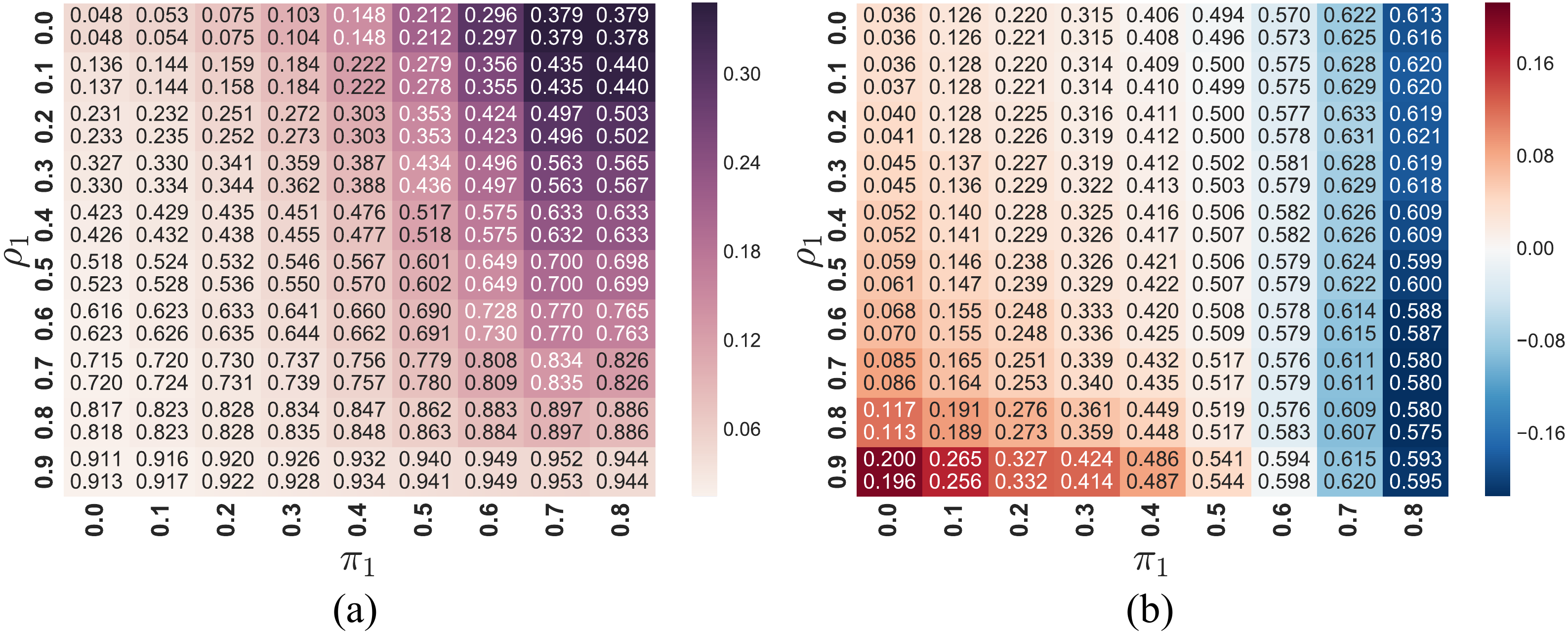}}
\vskip -0.08in
\caption{Rank Pruning $\hat{\rho}_1$ and $\hat{\pi}_1$ estimation consistency, averaged over all digits in MNIST. \textbf{(a)} Color depicts $\hat{\rho}_1-\rho_1$ with $\hat{\rho}_1$ (upper) and theoretical $\hat{\rho}_1^{thry}$ (lower) in each block. \textbf{(b)} Color depicts $\hat{\pi}_1-\pi_1$ with $\hat{\pi}_1$ (upper) and $\hat{\pi}_1^{thry}$ (lower) in each block. }
\label{rho1_pi1_mnist_logreg}
\end{center}
\vskip -0.1in
\end{figure} 

As evidence that Rank Pruning is dataset and classifier agnostic, we demonstrate its superiority with both (1) a linear logistic regression model with unit L2 regularization and (2) an AlexNet CNN variant with max pooling and dropout, modified to have a two-class output. The CNN structure is adapted from \cite{mnist_cnn_structure} for MNIST and \cite{cifar_cnn_structure} for CIFAR. CNN training ends when a 10\% holdout set shows no loss decrease for 10 epochs (max 50 for MNIST and 150 for CIFAR).

We consider noise rates \begin{small}{$\pi_1, \rho_1\in\{(0, 0.5),(0.25,0.25),$
$(0.5,0),(0.5,0.5)\}$}\end{small} for both MNIST and CIFAR, with additional settings for MNIST in Table \ref{table:mnist_cifar_logreg_f1} to emphasize Rank Pruning performance is noise rate agnostic. The $\rho_1=0$, $\pi_1=0$ case is omitted because when given $\rho_1$, $\pi_1$, all methods have the same loss function as the ground truth classifier, resulting in nearly identical F1 scores. Note that in general, Rank Pruning does not require perfect probability estimation to achieve perfect F1-score. As an example, this occurs when $P$ and $N$ are range-separable, and the rank order of the sorted $g(x)$ probabilities in $P$ and $N$ is consistent with the rank of the perfect probabilities, regardless of the actual values of $g(x)$.

For MNIST using logistic regression, we evaluate the consistency of our noise rate estimates with actual noise rates and theoretical estimates (Eq. \ref{rho_imperfect_condition}) across $\pi_1\in[0, 0.8] \times \rho_1\in[0,0.9]$. The computing time for one setting was $\sim10$ minutes on a single CPU core. The results for $\hat{\rho}_1$ and $\hat{\pi}_1$ (Fig. \ref{rho1_pi1_mnist_logreg}) are satisfyingly consistent, with mean absolute difference MD$_{\hat{\rho}_1, \rho_1}=0.105$ and  MD$_{\hat{\pi}_1, \pi_1}=0.062$, and validate our theoretical solutions (MD$_{\hat{\rho}_1,\hat{\rho}_1^{thry}}=0.0028$, MD$_{\hat{\pi}_1,\hat{\pi}_1^{thry}}=0.0058$). The deviation of the theoretical and empirical estimates reflects the assumption that we have infinite examples, whereas empirically, the number of examples is finite.

We emphasize two observations from our analysis on CIFAR and MNIST. First, Rank Pruning performs well in nearly every scenario and boasts the most dramatic improvement over prior state-of-the-art in the presence of extreme noise ($\pi_1=0.5$, $\rho_1=0.5$). This is easily observed in the right-most quadrant of Table \ref{table:mnist_cifar_cnn_f1}. The $\pi_1=0.5$, $\rho_1=0$ quadrant is nearest to $\pi_1=0$, $\rho_1=0$ and mostly captures CNN prediction variation because $|\tilde{P}| \ll |\tilde{N}|$.

Second, RP$_{\rho}$ often achieves equivalent (MNIST in Table \ref{table:mnist_cifar_cnn_f1}) or significantly higher (CIFAR in Tables \ref{table:mnist_cifar_logreg_f1} and \ref{table:mnist_cifar_cnn_f1}) F1 score than Rank Pruning when $\rho_1$ and $\pi_1$ are provided, particularly when noise rates are large. This effect is exacerbated for harder problems (lower F1 score for the ground truth classifier) like the ``cat” in CIFAR or the ``9” digit in MNIST likely because these problems are more complex, resulting in less confident predictions, and therefore more pruning. 

Remember that $\rho^{conf}_1$ and $\rho^{conf}_0$ are upper bounds when $g$ is unassuming. Noise rate overestimation accounts for the complexity of harder problems. As a downside, Rank Pruning may remove correctly labeled examples that ``confuse” the classifier, instead fitting only the confident examples in each class. We observe this on CIFAR in Table \ref{table:mnist_cifar_logreg_f1} where logistic regression severely underfits so that RP$_{\rho}$ has significantly higher F1 score than the ground truth classifier. Although Rank Pruning with noisy labels seemingly outperforms the ground truth model, if we lower the classification threshold to 0.3 instead of 0.5, the performance difference goes away by accounting for the lower probability predictions.

\begin{table*}[t]

\setlength\tabcolsep{2pt} 
\renewcommand{\arraystretch}{0.9}
\caption{Comparison of F1 score for one-vs-rest MNIST and CIFAR-10 (averaged over all digits/images) using logistic regression. Except for $RP_\rho$, $\rho_1$, $\rho_0$ are given to all methods. Top model scores are in bold with $RP_\rho$ in red if greater than non-RP models. Due to sensitivity to imperfect $g(x)$, \emph{Liu16} often
predicts the same label for all examples.
} 
\vskip -0.17in
\label{table:mnist_cifar_logreg_f1}
\begin{center}
\begin{small}
\begin{sc}

\resizebox{\textwidth}{!}{ 
\begin{tabular}{l|cccc|ccc|cccc|cccc|cccc}
\toprule

\multicolumn{0}{r|}{\textbf{Dataset}} & \multicolumn{4}{c|}{\textbf{CIFAR}} & \multicolumn{15}{c}{\textbf{MNIST}} \\

\multicolumn{0}{r|}{$\pi_1=$} & \textbf{0.0} & \textbf{0.25} & \textbf{0.5} & \textbf{0.5} &
\multicolumn{3}{c}{$\pi_1$\textbf{ = 0.0}}  &   
\multicolumn{4}{c}{$\pi_1$\textbf{ = 0.25}}  & 
\multicolumn{4}{c}{$\pi_1$\textbf{ = 0.5}}  & 
\multicolumn{4}{c}{$\pi_1$\textbf{ = 0.75}} 
\\

\textbf{Model,$\rho_1 = $} &   \textbf{0.5} &    \textbf{0.25} &    \textbf{0.0} &   \textbf{0.5} &  \textbf{0.25} &    \textbf{0.5} &   \textbf{0.75} &   \textbf{0.0} &    \textbf{0.25} &    \textbf{0.5} &   \textbf{0.75} &   \textbf{0.0} &    \textbf{0.25} &    \textbf{0.5} &   \textbf{0.75} &   \textbf{0.0} &    \textbf{0.25} &    \textbf{0.5} &   \textbf{0.75}    \\

\midrule

\textbf{True}   &  0.248 & 0.248 & 0.248 & 0.248 & 0.894 &  0.894 &  0.894 &  0.894 &  0.894 &  0.894 &  0.894 &  0.894 &  0.894 &  0.894 &  0.894 &  0.894 &  0.894 &  0.894 &  0.894 \\
\textbf{RP}$_{\rho}$ &  \textcolor{red}{\textbf{0.301}}  & \textcolor{red}{\textbf{0.316}}  & \textcolor{red}{\textbf{0.308}}  & \textcolor{red}{\textbf{0.261}}  &  \textcolor{red}{\textbf{0.883}} &  \textcolor{red}{\textbf{0.874}} &  \textcolor{red}{\textbf{0.843}} &  \textcolor{red}{\textbf{0.881}} &  \textcolor{red}{\textbf{0.876}} &  \textcolor{red}{\textbf{0.863}} &  \textcolor{red}{\textbf{0.799}} &  0.823 &  0.831 &  \textcolor{red}{\textbf{0.819}} &  \textcolor{red}{\textbf{0.762}} &  0.583 &  0.603 &  0.587 &  0.532 \\
\textbf{RP} & \textbf{0.256} & \textbf{0.262} & \textbf{0.244} & 0.209 & \textbf{0.885} &  \textbf{0.873} &  \textbf{0.839} &  \textbf{0.890} &  \textbf{0.879} &  \textbf{0.863} &  \textbf{0.812} &  \textbf{0.879} &  \textbf{0.862} &  \textbf{0.838} &  \textbf{0.770} &  \textbf{0.855} &  \textbf{0.814} &  \textbf{0.766} &  0.617 \\
\textbf{Nat13} & 0.226 & 0.219 & 0.194 & 0.195 &  0.860 &  0.830 &  0.774 &  0.865 &  0.836 &  0.802 &  0.748 &  0.839 &  0.810 &  0.777 &  0.721 &  0.809 &  0.776 &  0.736 &  \textbf{0.640} \\
\textbf{Elk08} & 0.221  & 0.226  & 0.228  & \textbf{0.210}  &  0.862 &  0.830 &  0.771 &  0.864 &  0.847 &  0.819 &  0.762 &  0.843 &  0.835 &  0.814 &  0.736 &  0.674 &  0.669 &  0.599 &  0.473 \\
\textbf{Liu16} & 0.182 & 0.182 & 0.000 & 0.182 &   0.021 &  0.000 &  0.000 &  0.000 &  0.147 &  0.147 &  0.073 &  0.000 &  0.164 &  0.163 &  0.163 &  0.047 &  0.158 &  0.145 &  0.164 \\

\bottomrule
\end{tabular}
}
\end{sc}
\end{small}
\end{center}
\vskip -0.1in
\end{table*}


\begin{table*}[t]

\setlength\tabcolsep{1pt} 
\renewcommand{\arraystretch}{0.85}
\caption{F1 score comparison on MNIST and CIFAR-10 using a CNN. Except for $RP_\rho$, $\rho_1$, $\rho_0$ are given to all methods.
} 
\vskip -0.17in
\label{table:mnist_cifar_cnn_f1}
\begin{center}
\begin{small}
\begin{sc}

\resizebox{\textwidth}{!}{
\begin{tabular}{l|c|ccccc|ccccc|ccccc|ccccc}
\toprule

\multicolumn{2}{l|}{\textbf{MNIST/CIFAR}} &  
\multicolumn{5}{c|}{$\pi_1$\textbf{ = 0.0}}   &   
\multicolumn{5}{c|}{$\pi_1$\textbf{ = 0.25}}  & 
\multicolumn{10}{c}{$\pi_1$\textbf{ = 0.5}}   \\

\multicolumn{1}{l}{\textbf{IMAGE}} &  
\multicolumn{1}{c|}{} & 
\multicolumn{5}{c|}{$\rho_1$\textbf{ = 0.5}} & 
\multicolumn{5}{c|}{$\rho_1$\textbf{ = 0.25}}  & 
\multicolumn{5}{c}{$\rho_1$\textbf{ = 0.0}} & 
\multicolumn{5}{c}{$\rho_1$\textbf{ = 0.5}}    \\

{\textbf{CLASS}} &   \textbf{True} & \textbf{RP}$_{\rho}$ &    \textbf{RP} & \textbf{Nat13} & \textbf{Elk08} & \textbf{Liu16} & \textbf{RP}$_{\rho}$ &    \textbf{RP} & \textbf{Nat13} & \textbf{Elk08} & \textbf{Liu16} & \textbf{RP}$_{\rho}$ &    \textbf{RP} & \textbf{Nat13} & \textbf{Elk08} & \textbf{Liu16} & \textbf{RP}$_{\rho}$ &    \textbf{RP} & \textbf{Nat13} & \textbf{Elk08} & \textbf{Liu16} \\
\midrule
\textbf{0}     &  0.993 &  \textcolor{red}{\textbf{0.991}} &  \textbf{0.988} &  0.977 &  0.976 &  0.179 &  \textcolor{red}{\textbf{0.991}} &  \textbf{0.992} &  0.982 &  0.981 &  0.179 &  \textcolor{red}{\textbf{0.991}} &  \textbf{0.992} &  0.984 &  0.987 &  0.985 &  \textcolor{red}{\textbf{0.989}} &  \textbf{0.989} &  0.937 &  0.964 &  0.179 \\
\textbf{1}     &  0.993 &  \textcolor{red}{\textbf{0.990}} &  \textbf{0.991} &  0.989 &  0.985 &  0.204 &  \textcolor{red}{\textbf{0.992}} &  \textbf{0.992} &  0.984 &  0.987 &  0.204 &  0.990 &  0.991 &  0.992 &  \textbf{0.993} &  0.990 &  \textcolor{red}{\textbf{0.989}} &  \textbf{0.989} &  0.984 &  0.988 &  0.204 \\
\textbf{2}     &  0.987 &  \textcolor{red}{\textbf{0.973}} &  \textbf{0.976} &  0.972 &  0.969 &  0.187 &  \textcolor{red}{\textbf{0.984}} &  \textbf{0.983} &  0.978 &  0.975 &  0.187 &  0.985 &  0.986 &  0.985 &  0.986 &  \textbf{0.988} &  \textcolor{red}{\textbf{0.971}} &  \textbf{0.975} &  0.968 &  0.959 &  0.187 \\
\textbf{3}     &  0.990 &  \textcolor{red}{\textbf{0.984}} &  \textbf{0.984} &  0.972 &  0.981 &  0.183 &  \textcolor{red}{\textbf{0.986}} &  \textbf{0.986} &  0.978 &  0.978 &  0.183 &  \textcolor{red}{\textbf{0.990}} &  0.987 &  \textbf{0.989} &  \textbf{0.989} &  0.984 &  \textcolor{red}{\textbf{0.981}} &  \textbf{0.979} &  0.957 &  0.971 &  0.183 \\
\textbf{4}     &  0.994 &  \textcolor{red}{\textbf{0.981}} &  0.979 &  \textbf{0.981} &  0.977 &  0.179 &  \textcolor{red}{\textbf{0.985}} &  \textbf{0.987} &  0.971 &  0.964 &  0.179 &  0.987 &  \textbf{0.990} &  \textbf{0.990} &  0.989 &  0.985 &  \textcolor{red}{\textbf{0.977}} &  \textbf{0.982} &  0.955 &  0.961 &  0.179 \\
\textbf{5}     &  0.989 &  \textcolor{red}{\textbf{0.982}} &  \textbf{0.980} &  0.978 &  0.979 &  0.164 &  \textcolor{red}{\textbf{0.985}} &  \textbf{0.982} &  0.964 &  0.965 &  0.164 &  \textcolor{red}{\textbf{0.988}} &  \textbf{0.987} &  \textbf{0.987} &  0.984 &  \textbf{0.987} &  \textcolor{red}{\textbf{0.965}} &  \textbf{0.968} &  0.962 &  0.957 &  0.164 \\
\textbf{6}     &  0.989 &  \textcolor{red}{\textbf{0.986}} &  \textbf{0.985} &  0.972 &  0.982 &  0.175 &  \textcolor{red}{\textbf{0.985}} &  \textbf{0.987} &  0.978 &  0.981 &  0.175 &  0.985 &  0.985 &  \textbf{0.988} &  0.987 &  0.985 &  \textcolor{red}{\textbf{0.983}} &  \textbf{0.982} &  0.946 &  0.959 &  0.175 \\
\textbf{7}     &  0.987 &  \textcolor{red}{\textbf{0.981}} &  \textbf{0.980} &  0.967 &  0.948 &  0.186 &  \textcolor{red}{\textbf{0.976}} &  \textbf{0.975} &  0.971 &  0.971 &  0.186 &  0.976 &  0.980 &  \textbf{0.985} &  0.982 &  0.983 &  \textcolor{red}{\textbf{0.973}} &  \textbf{0.968} &  0.942 &  0.958 &  0.186 \\
\textbf{8}     &  0.989 &  \textcolor{red}{\textbf{0.975}} &  \textbf{0.978} &  0.943 &  0.967 &  0.178 &  \textcolor{red}{\textbf{0.982}} &  \textbf{0.981} &  0.967 &  0.951 &  0.178 &  0.982 &  \textbf{0.984} &  0.982 &  0.979 &  0.983 &  \textcolor{red}{\textbf{0.977}} &  \textbf{0.975} &  0.864 &  0.959 &  0.178 \\
\textbf{9}     &  0.982 &  0.966 &  \textbf{0.974} &  0.972 &  0.935 &  0.183 &  \textcolor{red}{\textbf{0.976}} &  \textbf{0.974} &  0.967 &  0.967 &  0.183 &  0.976 &  0.975 &  0.974 &  \textbf{0.978} &  0.970 &  \textcolor{red}{\textbf{0.959}} &  0.940 &  0.931 &  \textbf{0.942} &  0.183 \\
\midrule
\textbf{AVG$_{MN}$} &  0.989 &  \textcolor{red}{\textbf{0.981}} &  \textbf{0.981} &  0.972 &  0.970 &  0.182 &  \textcolor{red}{\textbf{0.984}} &  \textbf{0.984} &  0.974 &  0.972 &  0.182 &  0.985 &  \textbf{0.986} &  \textbf{0.986} &  0.985 &  0.984 &  \textcolor{red}{\textbf{0.976}} &  \textbf{0.975} &  0.945 &  0.962 &  0.182 \\

\midrule

\textbf{plane}   &  0.755 &  \textcolor{red}{\textbf{0.689}} &  \textbf{0.634} &  0.619 &  0.585 &  0.182 &  \textcolor{red}{\textbf{0.695}} &  \textbf{0.702} &  0.671 &  0.640 &  0.182 &  \textcolor{red}{\textbf{0.757}} &  \textbf{0.746} &  0.716 &  0.735 &    0.000 &  \textcolor{red}{\textbf{0.628}} &  \textbf{0.635} &  0.459 &  0.598 &  0.182 \\
\textbf{auto} &  0.891 &  \textcolor{red}{\textbf{0.791}} &  \textbf{0.785} &  0.761 &  0.768 &  0.000 &  \textcolor{red}{\textbf{0.832}} &  \textbf{0.824} &  0.771 &  0.783 &  0.182 &  0.862 &  0.866 &  \textbf{0.869} &  0.865 &    0.000 &  \textcolor{red}{\textbf{0.749}} &  \textbf{0.720} &  0.582 &  0.501 &  0.182 \\
\textbf{bird}       &  0.669 &  \textcolor{red}{\textbf{0.504}} &  \textbf{0.483} &  0.445 &  0.389 &  0.182 &  \textcolor{red}{\textbf{0.543}} &  \textbf{0.515} &  0.469 &  0.426 &  0.182 &  \textcolor{red}{\textbf{0.577}} &  \textbf{0.619} &  0.543 &  0.551 &    0.000 &  \textcolor{red}{\textbf{0.447}} &  \textbf{0.409} &  0.366 &  0.387 &  0.182 \\
\textbf{cat}        &  0.487 &  \textcolor{red}{\textbf{0.350}} &  0.279 &  0.310 &  \textbf{0.313} &  0.000 &  \textcolor{red}{\textbf{0.426}} &  0.317 &  \textbf{0.350} &  0.345 &  0.182 &  \textcolor{red}{\textbf{0.489}} &  \textbf{0.433} &  0.426 &  0.347 &    0.000 &  \textcolor{red}{\textbf{0.394}} &  0.282 &  0.240 &  \textbf{0.313} &  0.182 \\
\textbf{deer}       &  0.726 &  \textcolor{red}{\textbf{0.593}} &  \textbf{0.540} &  0.455 &  0.522 &  0.182 &  \textcolor{red}{\textbf{0.585}} &  0.554 &  0.480 &  \textbf{0.569} &  0.182 &  0.614 &  0.630 &  \textbf{0.643} &  0.633 &    0.000 &  \textcolor{red}{\textbf{0.458}} &  0.375 &  0.310 &  \textbf{0.383} &  0.182 \\
\textbf{dog}        &  0.569 &  \textcolor{red}{\textbf{0.544}} &  \textbf{0.577} &  0.429 &  0.456 &  0.000 &  \textcolor{red}{\textbf{0.579}} &  0.559 &  0.569 &  \textbf{0.576} &  0.182 &  0.647 &  0.637 &  \textbf{0.667} &  0.630 &    0.000 &  \textcolor{red}{\textbf{0.516}} &  0.461 &  0.412 &  \textbf{0.465} &  0.182 \\
\textbf{frog}       &  0.815 &  \textcolor{red}{\textbf{0.746}} &  0.727 &  \textbf{0.733} &  0.718 &  0.000 &  \textcolor{red}{\textbf{0.729}} &  \textbf{0.750} &  0.630 &  0.584 &  0.182 &  0.767 &  \textbf{0.782} &  0.777 &  0.770 &    0.000 &  \textcolor{red}{\textbf{0.635}} &  \textbf{0.615} &  0.589 &  0.524 &  0.182 \\
\textbf{horse}      &  0.805 &  \textcolor{red}{\textbf{0.690}} &  0.670 &  0.624 &  \textbf{0.672} &  0.182 &  \textcolor{red}{\textbf{0.710}} &  0.669 &  \textbf{0.683} &  0.627 &  0.182 &  0.761 &  \textbf{0.776} &  0.769 &  0.753 &    0.000 &  \textcolor{red}{\textbf{0.672}} &  \textbf{0.569} &  0.551 &  0.461 &  0.182 \\
\textbf{ship}       &  0.851 &  \textcolor{red}{\textbf{0.791}} &  \textbf{0.783} &  0.719 &  0.758 &  0.182 &  \textcolor{red}{\textbf{0.810}} &  \textbf{0.801} &  0.758 &  0.723 &  0.182 &  0.816 &  0.822 &  0.830 &  \textbf{0.831} &    0.000 &  \textcolor{red}{\textbf{0.715}} &  \textbf{0.738} &  0.569 &  0.632 &  0.182 \\
\textbf{truck}      &  0.861 &  \textcolor{red}{\textbf{0.744}} &  \textbf{0.722} &  0.655 &  0.665 &  0.182 &  \textcolor{red}{\textbf{0.814}} &  \textbf{0.826} &  0.798 &  0.774 &  0.182 &  0.812 &  0.830 &  \textbf{0.826} &  0.824 &    0.000 &  \textcolor{red}{\textbf{0.654}} &  0.543 &  0.575 &  \textbf{0.584} &  0.182 \\
\midrule
\textbf{AVG$_{CF}$}      &  0.743 &  \textcolor{red}{\textbf{0.644}} &  \textbf{0.620} &  0.575 &  0.585 &  0.109 &  \textcolor{red}{\textbf{0.672}} &  \textbf{0.652} &  0.618 &  0.605 &  0.182 &  \textcolor{red}{\textbf{0.710}} &  \textbf{0.714} &  0.707 &  0.694 &    0.000 &  \textcolor{red}{\textbf{0.587}} &  \textbf{0.535} &  0.465 &  0.485 &  0.182 \\
\bottomrule
\end{tabular}
}
\end{sc}
\end{small}
\end{center}
\vskip -0.1in
\end{table*}
\section{Discussion}

To our knowledge, Rank Pruning is the first time-efficient algorithm, w.r.t. classifier fitting time, for $\tilde{P}\tilde{N}$ learning that achieves similar or better F1, error, and AUC-PR than current state-of-the-art methods across practical scenarios for synthetic, MNIST, and CIFAR datasets, with logistic regression and CNN classifiers, across all noise rates, $\rho_1, \rho_0$, for varying added noise, dimension, separability, and number of training examples. By \emph{learning with confident examples}, we discover provably consistent estimators for noise rates, $\rho_1$, $\rho_0$, derive theoretical solutions when $g$ is unassuming, and accurately uncover the classifications of $f$ fit to hidden labels, perfectly when $g$ range separates $P$ and $N$.

We recognize that disambiguating whether we are in the unassuming or range separability condition may be desirable. Although knowing $g^*(x)$ and thus $\Delta g(x)$ is impossible, if we assume randomly uniform noise, and toggling the $LB_{y=1}$ threshold does not change $\rho^{conf}_1$, then $g$ range separates $P$ and $N$. When $g$ is unassuming, Rank Pruning is still robust to imperfect $g(x)$ within a range separable subset of $P$ and $N$ by training with confident examples even when noise rate estimates are inexact.

An important contribution of Rank Pruning is generality, both in classifier and implementation. The use of logistic regression and a generic CNN in our experiments emphasizes that our findings are not dependent on model complexity. We evaluate thousands of scenarios to avoid findings that are an artifact of problem setup. A key point of Rank Pruning is that we only report the simplest, non-parametric version. For example, we use 3-fold cross-validation to compute $g(x)$ even though we achieved improved performance with larger folds. We tried many variants of pruning and achieved significant higher F1 for MNIST and CIFAR, but to maintain generality, we present only the basic model.

At its core, Rank Pruning is a simple, robust, and general solution for noisy binary classification by \emph{learning with confident examples}, but it also challenges how we think about training data. For example, SVM showed how a decision boundary can be recovered from only support vectors. Yet, when training data contains significant mislabeling, confident examples, many of which are far from the boundary, are informative for uncovering the true relationship $P(y=1|x)$. Although modern affordances of ``big data" emphasize the value of \emph{more} examples for training, through Rank Pruning we instead encourage a rethinking of learning with \emph{confident} examples.

\newpage

\bibliography{rp}
\bibliographystyle{uai2017}

\newpage
\appendix
\numberwithin{equation}{section}

\quad

\begin{center}
\begin{huge}
\textbf{Appendix}
\end{huge}
\end{center}

\section{Proofs}
In this section, we provide proofs for all the lemmas and theorems in the main paper. We always assume that a class-conditional extension of the Classification Noise Process (CNP) \citep{angluin1988learning} maps true labels $y$ to observed labels $s$ such that each label in $P$ is flipped independently with probability $\rho_1$ and each label in $N$ is flipped independently with probability $\rho_0$ ($s \leftarrow CNP(y, \rho_1, \rho_0)$), so that $P(s=s|y=y,x) = P(s=s|y=y)$. Remember that $\rho_1 + \rho_0 <1$ is a necessary condition of minimal information, other we may learn opposite labels.

In Lemma \ref{thm:lemma1_appendix}, Theorem \ref{thm:theorem2_appendix}, Lemma \ref{thm:lemma3_appendix} and Theorem \ref{thm:theorem4_appendix}, we assume that $P$ and $N$ have infinite number of examples so that they are the true,  hidden distributions.

A fundamental equation we use in the proofs is the following lemma:
\medskip

\textbf{Lemma A1}
\textit{When $g$ is ideal, i.e. $g(x)=g^*(x)$ and $P$ and $N$ have non-overlapping support, we have}
\begin{equation} \label{eqA1}
\begin{split}
g(x) = (1 - \rho_1) \cdot \indicator{y=1} + \rho_0 \cdot \indicator{y=0}
\end{split}
\end{equation}

\textbf{Proof:}
Since $g(x)=g^*(x)$ and $P$ and $N$ have non-overlapping support, we have
\begin{equation*}
\begin{split}
g(x) = & g^*(x) = P(s=1|x) \\ 
 = & P(s=1|y=1,x) \cdot P(y=1|x) + P(s=1|y=0,x) \cdot P(y=0|x) \\
 = & P(s=1|y=1) \cdot P(y=1|x) + P(s=1|y=0) \cdot P(y=0|x) \\
 = & (1 - \rho_1) \cdot \mathbbm{1}[[y=1]] + \rho_0 \cdot \mathbbm{1}[[y=0]]
\end{split}
\end{equation*}

\subsection{Proof of Lemma 1}\label{sec:A1}

\textbf{Lemma \customlabel{thm:lemma1_appendix}{1}}
\textit{When $g$ is ideal, i.e. $g(x) = g^*(x)$ and $P$ and $N$ have non-overlapping support, we have}
\begin{equation}
\begin{cases}
\tilde{P}_{y=1} = \{x\in P|s=1\}, \tilde{N}_{y=1} = \{x\in P|s=0\}\\
\tilde{P}_{y=0} = \{x\in N|s=1\}, \tilde{N}_{y=0} = \{x\in N|s=0\}
\end{cases}
\end{equation}

\textbf{Proof:} Firstly, we compute the threshold $LB_{y=1}$ and $UB_{y=0}$ used by $\tilde{P}_{y=1}$, $\tilde{N}_{y=1}$, $\tilde{P}_{y=0} $ and $\tilde{N}_{y=0}$. Since $P$ and $N$ have non-overlapping support, we have $P(y=1|x)=\indicator{y=1}$. Also using $g(x) = g^*(x)$, we have

\vskip -0.2in

\begin{align*}
LB_{y=1}=&E_{x\in\tilde{P}}[g(x)]=E_{x\in\tilde{P}}[P(s=1|x)]\\
=&E_{x\in\tilde{P}}[P(s=1|x,y=1)P(y=1|x) +P(s=1|x,y=0)P(y=0|x)]\\
=&E_{x\in\tilde{P}}[P(s=1|y=1)P(y=1|x) +P(s=1|y=0)P(y=0|x)]\\
=&(1-\rho_1)(1-\pi_1)+\rho_0\pi_1\numberthis\label{rh1_prob_eq_A1}
\end{align*}

Similarly, we have
\begin{equation*}
UB_{y=0}=(1-\rho_1)\pi_0+\rho_0(1-\pi_0)
\end{equation*}

Since $\pi_1=P(y=0|s=1)$, we have $\pi_1\in[0,1]$. Furthermore, we have the requirement that $\rho_1+\rho_0<1$, then $\pi_1=1$ will lead to $\rho_1=P(s=0|y=1)=1-P(s=1|y=1)=1-\frac{P(y=1|s=1)P(s=1)}{P(y=1)}=1-0=1$ which violates the requirement of $\rho_1+\rho_0<1$. Therefore, $\pi_1\in[0,1)$. Similarly, we can prove $\pi_0\in[0,1)$. Therefore, we see that both $LB_{y=1}$ and $UB_{y=0}$ are interpolations of $(1-\rho_1)$ and $\rho_0$:
\begin{align*}
&\rho_0<LB_{y=1}\leq1-\rho_1\\
&\rho_0\leq UB_{y=0}<1-\rho_1
\end{align*}
The first equality holds iff $\pi_1=0$ and the second equality holds iff $\pi_0=0$.

Using Lemma A1, we know that under the condition of $g(x)=g^*(x)$ and non-overlapping support, $g(x)=(1 - \rho_1) \cdot \mathbbm{1}[[y=1]] + \rho_0 \cdot \mathbbm{1}[[y=0]]$. In other words, 
\begin{align*}
g(x)&\geq LB_{y=1}\Leftrightarrow x\in P\\
g(x)&\leq UB_{y=0}\Leftrightarrow x\in N
\end{align*}

Since
\begin{equation*}
\begin{cases}
\tilde{P}_{y=1}=\{x\in\tilde{P}|g(x)\geq LB_{y=1}\}\\
\tilde{N}_{y=1}=\{x\in\tilde{N}|g(x)\geq LB_{y=1}\}\\
\tilde{P}_{y=0}=\{x\in\tilde{P}|g(x)\leq UB_{y=0}\}\\
\tilde{N}_{y=0}=\{x\in\tilde{N}|g(x)\leq UB_{y=0}\}
\end{cases}
\end{equation*}

where $\tilde{P}=\{x|s=1\}$ and $\tilde{N}=\{x|s=0\}$, we have
\begin{equation*}
\begin{cases}
\tilde{P}_{y=1} = \{x\in P|s=1\}, \tilde{N}_{y=1} = \{x\in P|s=0\}\\
\tilde{P}_{y=0} = \{x\in N|s=1\}, \tilde{N}_{y=0} = \{x\in N|s=0\}
\end{cases}
\end{equation*}

\subsection{Proof of Theorem 2}
We restate Theorem 2 here:

\textbf{Theorem \customlabel{thm:theorem2_appendix}{2}}
\textit{When $g$ is ideal, i.e. $g(x) = g^*(x)$ and $P$ and $N$ have non-overlapping support, we have}

\vskip -0.2in

\begin{align*}
\hat{\rho}_1^{conf}=\rho_1,\hat{\rho}_0^{conf}=\rho_0
\end{align*}

\textbf{Proof:} 
Using the definition of $\hat{\rho}_1^{conf}$ in the main paper:

\begin{equation*}
\hat{\rho}_1^{conf}=\frac{|\tilde{N}_{y=1}|}{|\tilde{N}_{y=1}|+|\tilde{P}_{y=1}|},\ 
\hat{\rho}_0^{conf}=\frac{|\tilde{P}_{y=0}|}{|\tilde{P}_{y=0}|+|\tilde{N}_{y=0}|}
\end{equation*}

Since $g(x) = g^*(x)$ and $P$ and $N$ have non-overlapping support, using Lemma 1, we know 
\begin{equation*}
\begin{cases}
\tilde{P}_{y=1} = \{x\in P|s=1\}, \tilde{N}_{y=1} = \{x\in P|s=0\}\\
\tilde{P}_{y=0} = \{x\in N|s=1\}, \tilde{N}_{y=0} = \{x\in N|s=0\}
\end{cases}
\end{equation*}
Since $\rho_1=P(s=0|y=1)$ and $\rho_0=P(s=1|y=0)$, we immediately have
\begin{align*}
\hat{\rho}_1^{conf}=\frac{|\{x\in P|s=0\}|}{|P|}=\rho_1,\ \hat{\rho}_0^{conf}=\frac{|\{x\in N|s=1\}|}{|N|}=\rho_0
\end{align*}

\subsection{Proof of Lemma 3}
We rewrite Lemma 3 below:

\textbf{Lemma \customlabel{thm:lemma3_appendix}{3}}
\textit{When $g$ is unassuming, i.e., $\Delta g(x):=g(x)-g^*(x)$ can be nonzero, and $P$ and $N$ can have overlapping support, we have}

\vskip -0.2in

\begin{equation}\label{rho_imperfect_condition_appendix}
\begin{cases}
LB_{y=1}=LB_{y=1}^*+E_{x\in\tilde{P}}[\Delta g(x)]-\frac{(1-\rho_1-\rho_0)^2}{p_{s1}}\Delta p_o\\
UB_{y=0}=UB_{y=0}^*+E_{x\in\tilde{N}}[\Delta g(x)]+\frac{(1-\rho_1-\rho_0)^2}{1-p_{s1}}\Delta p_o\\
\hat{\rho}_1^{conf}=\rho_1+\frac{1-\rho_1-\rho_0}{|P|-|\Delta P_1| + |\Delta N_1|}|\Delta N_1|\\
\hat{\rho}_0^{conf}=\rho_0+\frac{1-\rho_1-\rho_0}{|N|-|\Delta N_0| + |\Delta P_0|}|\Delta P_0|\\
\end{cases}
\end{equation}

\vskip -0.1in
\textit{where}
\vskip -0.1in

\begin{equation}\label{def_all_delta_P_N}
\begin{cases}
LB_{y=1}^*=(1-\rho_1)(1-\pi_1)+\rho_0\pi_1\\ UB_{y=0}^*=(1-\rho_1)\pi_0+\rho_0(1-\pi_0)\\
\Delta p_o:=\frac{|P \cap N|}{|P \cup N|}\\
\Delta P_1=\{x\in P|g(x)< LB_{y=1}\}\\
\Delta N_1=\{x\in N|g(x)\geq LB_{y=1}\}\\
\Delta P_0=\{x\in P|g(x)\leq UB_{y=0}\}\\
\Delta N_0=\{x\in N|g(x)> UB_{y=0}\}\\
\end{cases}
\end{equation}


\medskip

\noindent \textbf{Proof:} We first calculate $LB_{y=1}$ and $UB_{y=0}$ under unassuming conditions, then calculate $\hat{\rho}_i^{conf}$, $i=0,1$ under unassuming condition.

Note that $\Delta p_o$ can also be expressed as
\begin{align*}
\Delta p_o:=\frac{|P \cap N|}{|P \cup N|}=P(\hat{y}=1,y=0)=P(\hat{y}=0,y=1)
\end{align*}

Here $P(\hat{y}=1,y=0)\equiv P(\hat{y}=1|y=0)P(y=0)$, where $P(\hat{y}=1|y=0)$ means for a perfect classifier $f^*(x)=P(y=1|x)$, the expected probability that it will label a $y=0$ example as positive ($\hat{y}=1$).

\medskip

\textbf{(1) $LB_{y=1}$ and $UB_{y=0}$ under unassuming condition}

Firstly, we calculate $LB_{y=1}$ and $UB_{y=0}$ with perfect probability estimation $g^*(x)$, but the support may overlap. Secondly, we allow the probability estimation to be imperfect, superimposed onto the overlapping support condition, and calculate $LB_{y=1}$ and $UB_{y=0}$.

\textbf{I. Calculating $LB_{y=1}$ and $UB_{y=0}$ when $g(x)=g^*(x)$ and support may overlap}

With overlapping support, we no longer have $P(y=1|x)=\indicator{y=1}$. Instead, we have




\vskip -0.2in

\begin{align*}
LB_{y=1}=&E_{x\in\tilde{P}}[g^*(x)]=E_{x\in\tilde{P}}[P(s=1|x)]\\
=&E_{x\in\tilde{P}}[P(s=1|x,y=1)P(y=1|x) +P(s=1|x,y=0)P(y=0|x)]\\
=&E_{x\in\tilde{P}}[P(s=1|y=1)P(y=1|x) +P(s=1|y=0)P(y=0|x)]\\
=&(1-\rho_1)\cdot E_{x\in\tilde{P}}[P(y=1|x)]+\rho_0\cdot E_{x\in\tilde{P}}[P(y=0|x)]\\
=&(1-\rho_1)\cdot P(\hat{y}=1|s=1)+\rho_0\cdot P(\hat{y}=0|s=1)\\
\end{align*}

\vskip -0.2in

Here $P(\hat{y}=1|s=1)$ can be calculated using $\Delta p_o$:

\vskip -0.15in

\begin{align*}
P(\hat{y}=1|s=1)&=\frac{P(\hat{y}=1,s=1)}{P(s=1)}\\
&=\frac{P(\hat{y}=1,y=1,s=1)+P(\hat{y}=1,y=0,s=1)}{P(s=1)}\\
&=\frac{P(s=1|y=1)P(\hat{y}=1,y=1)+P(s=1|y=0)P(\hat{y}=1,y=0)}{P(s=1)}\\
&=\frac{(1-\rho_1)(p_{y1}-\Delta p_o)+\rho_0\Delta p_o}{p_{s1}}\\
&=(1-\pi_1)-\frac{1-\rho_1-\rho_0}{p_{s1}}\Delta p_o
\end{align*}

Hence,
\begin{align*}
P(\hat{y}=0|s=1)=1-P(\hat{y}=1|s=1)=\pi_1+\frac{1-\rho_1-\rho_0}{p_{s1}}\Delta p_o
\end{align*}

Therefore,
\begin{align*}
LB_{y=1}&=(1-\rho_1)\cdot P(\hat{y}=1|s=1)+\rho_0\cdot P(\hat{y}=0|s=1)\\
&=(1-\rho_1)\cdot \left((1-\pi_1)-\frac{1-\rho_1-\rho_0}{p_{s1}}\Delta p_o\right)+\rho_0\cdot \left(\pi_1+\frac{1-\rho_1-\rho_0}{p_{s1}}\Delta p_o\right)\\
&=LB_{y=1}^*-\frac{(1-\rho_1-\rho_0)^2}{p_{s1}}\Delta p_o\numberthis \label{rho1_prob_overlap}
\end{align*}

where $LB_{y=1}^*$ is the $LB_{y=1}$ value when $g(x)$ is ideal. We see in Eq. (\ref{rho1_prob_overlap}) that the overlapping support introduces a non-positive correction to $LB_{y=1}^*$ compared with the ideal condition.

Similarly, we have
\begin{align}\label{rho0_prob_overlap}
UB_{y=0}=UB_{y=0}^*+\frac{(1-\rho_1-\rho_0)^2}{1-p_{s1}}\Delta p_o
\end{align}

\medskip

\textbf{II. Calculating $LB_{y=1}$ and $UB_{y=0}$ when $g$ is unassuming}

Define $\Delta g(x)=g(x)-g^*(x)$. When the support may overlap, we have

\begin{align*}
LB_{y=1}&=E_{x\in\tilde{P}}[g(x)]\\
&=E_{x\in\tilde{P}}[g^*(x)]+E_{x\in\tilde{P}}[\Delta g(x)]\\
&=LB_{y=1}^*-\frac{(1-\rho_1-\rho_0)^2}{p_{s1}}\Delta p_o+E_{x\in\tilde{P}}[\Delta g(x)]\numberthis\label{rho1_nonideal}
\end{align*}

Similarly, we have
\begin{align*}
UB_{y=0}&=E_{x\in\tilde{N}}[g(x)]\\
&=E_{x\in\tilde{N}}[g^*(x)]+E_{x\in\tilde{N}}[\Delta g(x)]\\
&=UB_{y=0}^*+\frac{(1-\rho_1-\rho_0)^2}{1-p_{s1}}\Delta p_o+E_{x\in\tilde{N}}[\Delta g(x)]\numberthis\label{rho0_nonideal}
\end{align*}

In summary, Eq. (\ref{rho1_nonideal}) (\ref{rho0_nonideal}) give the expressions for $LB_{y=1}$ and $UB_{y=0}$, respectively, when $g$ is unassuming.

\medskip

\textbf{(2) $\hat{\rho}_i^{conf}$ under unassuming condition}

Now let's calculate $\hat{\rho}_i^{conf}$, $i=0,1$. For simplicity, define
\begin{equation}\label{define_all_Delta}
\begin{cases}
PP=\{x\in P|s=1\}\\
PN=\{x\in P|s=0\}\\
NP=\{x\in N|s=1\}\\
NN=\{x\in N|s=0\}\\
\Delta_{PP_1}=\{x\in PP|g(x)< LB_{y=1}\}\\
\Delta_{NP_1}=\{x\in NP|g(x)\geq LB_{y=1}\}\\
\Delta_{PN_1}=\{x\in PN|g(x)< LB_{y=1}\}\\
\Delta_{NN_1}=\{x\in NN|g(x)\geq LB_{y=1}\}\\
\end{cases}
\end{equation}

For $\hat{\rho}_1^{conf}$, we have:
\begin{align*}
\hat{\rho}_1^{conf}=\frac{|\tilde{N}_{y=1}|}{|\tilde{P}_{y=1}| + |\tilde{N}_{y=1}|}
\end{align*}

Here
\begin{align*}
\tilde{P}_{y=1}&=\{x\in\tilde{P}|g(x)\geq LB_{y=1}\}\\
&=\{x\in PP|g(x)\geq LB_{y=1}\}\cup\{x\in NP|g(x)\geq LB_{y=1}\}\\
&=(PP\setminus \Delta_{PP_1})\cup\Delta_{NP_1}
\end{align*}

Similarly, we have
\begin{align*}
\tilde{N}_{y=1}=(PN\setminus\Delta_{PN_1})\cup\Delta_{NN_1}
\end{align*}

Therefore


\begin{align*}
\hat{\rho}_1^{conf}&=\frac{|PN|-|\Delta_{PN_1}|+|\Delta_{NN_1}|}{[(|PP|-|\Delta_{PP_1}|)+(|PN|-|\Delta_{PN_1}|)]+(|\Delta_{NN_1}|+|\Delta_{NP_1}|)}\\
&=\frac{|PN|-|\Delta_{PN_1}|+|\Delta_{NN_1}|}{|P|-|\Delta P_1|+|\Delta N_1|}\numberthis\label{rho_1_conf_general}
\end{align*}

where in the second equality we have used the definition of $\Delta P_1$ and $\Delta N_1$ in Eq. (\ref{def_all_delta_P_N}).

Using the definition of $\rho_1$, we have

\begin{align*}
\frac{|PN|-|\Delta_{PN_1}|}{|P|-|\Delta P_1|}&=\frac{|\{x\in PN|g(x)\geq LB_{y=1}\}|}{|\{x\in P|g(x)\geq  LB_{y=1}\}|}\\
&=\frac{P(x\in PN,g(x)\geq LB_{y=1})}{P(x\in P,g(x)\geq LB_{y=1})}\\
&=\frac{P(x\in PN|x\in P,g(x)\geq LB_{y=1})\cdot P(x\in P,g(x)\geq LB_{y=1})}{P(x\in P,g(x)\geq LB_{y=1})}\\
&=\frac{P(x\in PN|x\in P)\cdot P(x\in P,g(x)\geq LB_{y=1})}{P(x\in P,g(x)\geq LB_{y=1})}\\
&=\rho_1
\end{align*}
Here we have used the property of CNP that $(s \independent x) | y$, leading to $P(x\in PN|x\in P,g(x)\geq LB_{y=1})=P(x\in PN|x\in P)=\rho_1$.

Similarly, we have
\begin{align*}
\frac{|\Delta_{NN_1}|}{|\Delta N_1|}&= 1-\rho_0
\end{align*}

Combining with Eq. (\ref{rho_1_conf_general}), we have

\begin{align*}
\hat{\rho}_1^{conf}=\rho_1+ \frac{1-\rho_1-\rho_0}{|P|-|\Delta P_1| + |\Delta N_1|}|\Delta N_1|\numberthis\label{rho_1_conf_unassuming}
\end{align*}

Similarly, we have

\begin{align}\label{rho_0_conf_unassuming}
\hat{\rho}_0^{conf}=\rho_0+ \frac{1-\rho_1-\rho_0}{|N|-|\Delta N_0| + |\Delta P_0|}|\Delta P_0|
\end{align}

From the two equations above, we see that
\begin{equation}
\hat{\rho}_1^{conf}\geq \rho_1, \hat{\rho}_0^{conf}\geq \rho_0
\end{equation}

In other words, $\hat{\rho}_i^{conf}$ is an \textbf{\textit{upper bound}} of $\rho_i$, $i=0,1$. The equality for $\hat{\rho}_1^{conf}$ holds if $|\Delta N_1|=0$. The equality for $\hat{\rho}_0^{conf}$ holds if $|\Delta P_0|=0$.

\medskip

\subsection{Proof of Theorem 4}
Let's restate Theorem 4 below:

\textbf{Theorem \customlabel{thm:theorem4_appendix}{4}}
\textit{Given non-overlapping support condition,}

If $\forall x\in N, \Delta g(x)<LB_{y=1}-\rho_0$, then $\hat{\rho}_1^{conf}=\rho_1$.

If $\forall x\in P, \Delta g(x)>-(1-\rho_1-UB_{y=0}$), then $\hat{\rho}_0^{conf}=\rho_0$.

\medskip

Theorem 4 directly follows from Eq. (\ref{rho_1_conf_unassuming}) and (\ref{rho_0_conf_unassuming}). Assuming non-overlapping support, we have $g^*(x)=P(s=1|x)=(1-\rho_1)\cdot \indicator{y=1}+\rho_0\cdot \indicator{y=0}$. In other words, the contribution of overlapping support to $|\Delta N_1|$ and $|\Delta P_0|$ is 0.  The only source of deviation comes from imperfect $g(x)$.

For the first half of the theorem, since $\forall x\in N, \Delta g(x)<LB_{y=1}-\rho_0$, we have $\forall x\in N, g(x)=\Delta g(x)+g^*(x)<(LB_{y=1}-\rho_0)+\rho_0=LB_{y=1}$, then $|\Delta N_1|=|\{x\in N|g(x)\geq LB_{y=1}\}|=0$, so we have $\hat{\rho}_1^{conf}=\rho_1$. 

Similarly, for the second half of the theorem, since $\forall x\in P,  \Delta g(x)>-(1-\rho_1-UB_{y=0})$, then $|\Delta P_0|=|\{x\in P|g(x)\leq UB_{y=0}\}|=0$, so we have $\hat{\rho}_0^{conf}=\rho_0$.

\subsection{Proof of Theorem 5}\label{sec:A5}
Theorem 5 reads as follows:

\textbf{Theorem \customlabel{thm:theorem5_appendix}{5}}
\textit{If $g$ range separates $P$ and $N$ and $\hat{\rho}_i=\rho_i$, $i=0,1$, then for any classifier $f_{\theta}$ and any bounded loss function $l(\hat{y}_i,y_i)$, we have}

\begin{equation}
R_{\tilde{l},\mathcal{D}_{\rho}}(f_{\theta})=R_{l,\mathcal{D}}(f_{\theta})
\end{equation}

\textit{where $\tilde{l}(\hat{y}_i,s_i)$ is Rank Pruning's loss function given by
}

\begin{align*}\label{rankpruning_loss_function_A}
\tilde{l}(\hat{y_i}, s_i)=\frac{1}{1-\hat{\rho}_1}l(\hat{y_i}, s_i)\cdot\indicator{x_i\in \tilde{P}_{conf}}
+&\frac{1}{1-\hat{\rho}_0}l(\hat{y_i}, s_i)\cdot\indicator{x_i\in \tilde{N}_{conf}}\numberthis  
\end{align*}

\textit{and $\tilde{P}_{conf}$ and $\tilde{N}_{conf}$ are given by}

\begin{equation}\label{define_P_N_conf}
\tilde{P}_{conf} := \{x \in \tilde{P} \mid g(x) \geq k_1 \},
\tilde{N}_{conf} := \{x \in \tilde{N} \mid g(x) \leq k_0 \}
\end{equation}

\textit{where $k_1$ is the $(\hat{\pi}_1|\tilde{P}|)^{th}$ smallest $g(x)$ for $x \in \tilde{P}$ and $k_0$ is the $(\hat{\pi}_0|\tilde{N}|)^{th}$ largest $g(x)$ for $x \in \tilde{N}$}

\vskip 0.1in
\textbf{Proof:}

Since $\tilde{P}$ and $\tilde{N}$ are constructed from $P$ and $N$ with noise rates $\pi_1$ and $\pi_0$ using the class-conditional extension of the Classification Noise Process \citep{angluin1988learning}, we have

\begin{equation}
\begin{cases}
P = PP\cup PN\\
N = NP\cup NN\\
\tilde{P} = PP\cup NP\\
\tilde{N} = PN\cup NN\\
\end{cases}
\end{equation}

where
\begin{equation}
\begin{cases}
PP=\{x\in P|s=1\}\\
PN=\{x\in P|s=0\}\\
NP=\{x\in N|s=1\}\\
NN=\{x\in N|s=0\}
\end{cases}
\end{equation}

satisfying
\begin{equation}
\begin{cases}
PP \sim PN \sim P\\
NP \sim NN \sim N\\
\frac{|NP|}{|\tilde{P}|}=\pi_1,\frac{|PP|}{|\tilde{P}|}=1-\pi_1\\
\frac{|PN|}{|\tilde{N}|}=\pi_0,\frac{|NN|}{|\tilde{N}|}=1-\pi_0\\
\frac{|PN|}{|P|}=\rho_1,\frac{|PP|}{|P|}=1-\rho_1\\
\frac{|NP|}{|N|}=\rho_0,\frac{|NN|}{|N|}=1-\rho_0\\
\end{cases}
\end{equation}

Here the $\sim$ means obeying the same distribution.

Since $g$ range separates $P$ and $N$, there exists a real number $z$ such that $\forall x_1\in P$ and $\forall x_0 \in N$, we have $g(x_1)>z>g(x_0)$. Since $P=PP\cup PN$, $N=NP\cup NN$, we have 
\begin{align*}
\forall x\in PP, g(x)>z;\ \forall x\in PN, g(x)>z; \\
\forall x\in NP, g(x)<z;\ \forall x\in NN, g(x)<z\numberthis\label{g_range_separate_PNNP}
\end{align*}

Since $\hat{\rho}_1=\rho_1$ and $\hat{\rho}_0=\rho_0$, we have 

\begin{equation}
\begin{cases}
\hat{\pi}_1=\frac{\hat{\rho}_0}{p_{s1}}\frac{1-p_{s1}-\hat{\rho}_1}{1-\hat{\rho}_1-\hat{\rho}_0}=\frac{\rho_0}{p_{s1}}\frac{1-p_{s1}-\rho_1}{1-\rho_1-\rho_0}=\pi_1\equiv\frac{\rho_0|N|}{|\tilde{P}|}\\
\hat{\pi}_0=\frac{\hat{\rho}_1}{1-p_{s1}}\frac{p_{s1}-\hat{\rho}_0}{1-\hat{\rho}_1-\hat{\rho}_0}=\frac{\rho_1}{1-p_{s1}}\frac{p_{s1}-\rho_0}{1-\rho_1-\rho_0}=\pi_0\equiv\frac{\rho_1|P|}{|\tilde{N}|}
\end{cases}
\end{equation}

Therefore, $\hat{\pi}_1|\tilde{P}|=\pi_1|\tilde{P}|=\rho_0|N|$, $\hat{\pi}_0|\tilde{N}|=\pi_0|\tilde{N}|=\rho_1|P|$. Using $\tilde{P}_{conf}$ and $\tilde{N}_{conf}$'s definition in Eq. (\ref{define_P_N_conf}), and $g(x)$'s property in Eq. (\ref{g_range_separate_PNNP}), we have

\begin{equation}
\tilde{P}_{conf} = PP\sim P, \tilde{N}_{conf} = NN \sim N
\end{equation}

Hence $P_{conf}$ and $N_{conf}$ can be seen as a uniform downsampling of $P$ and $N$, with a downsampling ratio of $(1-\rho_1)$ for $P$ and $(1-\rho_0)$ for $N$. Then according to Eq. (\ref{rankpruning_loss_function_A}), the loss function $\tilde{l}(\hat{y}_i,s_i)$ essentially sees a fraction of $(1-\rho_1)$ examples in $P$ and a fraction of $(1-\rho_0)$ examples in $N$, with a final reweighting to restore the class balance. Then for any classifier $f_{\theta}$ that maps $x\to \hat{y}$ and any bounded loss function $l(\hat{y}_i,y_i)$, we have
\begin{align*}
R_{\tilde{l},\mathcal{D}_{\rho}}(f_{\theta})&=E_{(x,s)\sim \mathcal{D}_{\rho}}[\tilde{l}(f_{\theta}(x),s)]\\
&=\frac{1}{1-\hat{\rho}_1}\cdot E_{(x,s)\sim \mathcal{D}_{\rho}}\left[l(f_{\theta}(x),s)\cdot\indicator{x\in \tilde{P}_{conf}}\right]+\frac{1}{1-\hat{\rho}_0}\cdot E_{(x,s)\sim \mathcal{D}_{\rho}}\left[l(f_{\theta}(x),s)\cdot\indicator{x\in \tilde{N}_{conf}}\right]\\
&=\frac{1}{1-\rho_1}\cdot E_{(x,s)\sim \mathcal{D}_{\rho}}\left[l(f_{\theta}(x),s)\cdot\indicator{x\in \tilde{P}_{conf}}\right]+\frac{1}{1-\rho_0}\cdot E_{(x,s)\sim \mathcal{D}_{\rho}}\left[l(f_{\theta}(x),s)\cdot\indicator{x\in \tilde{N}_{conf}}\right]\\
&=\frac{1}{1-\rho_1}\cdot E_{(x,s)\sim \mathcal{D}_{\rho}}\left[l(f_{\theta}(x),s)\cdot\indicator{x\in PP}\right]+\frac{1}{1-\rho_0}\cdot E_{(x,s)\sim \mathcal{D}_{\rho}}\left[l(f_{\theta}(x),s)\cdot\indicator{x\in NN}\right]\\
&=\frac{1}{1-\rho_1}\cdot (1-\rho_1)\cdot E_{(x,y)\sim \mathcal{D}}\left[l(f_{\theta}(x),y)\cdot\indicator{x\in P}\right]+\frac{1}{1-\rho_0}\cdot (1-\rho_0)\cdot E_{(x,y)\sim \mathcal{D}}\left[l(f_{\theta}(x),y)\cdot\indicator{x\in N}\right]\\
&=E_{(x,y)\sim \mathcal{D}}\left[l(f_{\theta}(x),y)\cdot\indicator{x\in P}+l(f_{\theta}(x),y)\cdot\indicator{x\in N}\right]\\
&=E_{(x,y)\sim \mathcal{D}}\left[l(f_{\theta}(x),y)\right]\\
&=R_{l,\mathcal{D}}(f_{\theta})
\end{align*}

Therefore, we see that the expected risk for Rank Pruning with corrupted labels, is exactly the same as the expected risk for the true labels, for any bounded loss function $l$ and classifier $f_{\theta}$. The reweighting ensures that after pruning, the two sets still remain unbiased w.r.t. to the true dataset.

Since the ideal condition is more strict than the range separability condition, we immediately have that when $g$ is ideal and $\hat{\rho}_i=\rho_i$, $i=0,1$, $R_{\tilde{l},\mathcal{D}_{\rho}}(f_{\theta})=R_{l,\mathcal{D}}(f_{\theta})$ for any $f_{\theta}$ and bounded loss function $l$.

\section{Additional Figures}


Figure B\ref{rankprune_mean_mnist} shows the average image for each digit for the problem “1” or “not 1” in MNIST with logistic regression and high noise ($\rho_1=0.5, \pi_1=0.5$). The number on the bottom and on the right counts the total number of examples (images). From the figure we see that Rank Pruning makes few mistakes, and when it does, the mistakes vary greatly in image from the typical digit.



\section{Additional Tables}\label{sec:tables_appendix}

Here we provide additional tables for the comparison of error, Precision-Recall AUC (AUC-PR, \cite{Davis:2006:RPR:1143844.1143874}), and F1 score for the algorithms \emph{RP}, \emph{Nat13}, \emph{Elk08}, \emph{Liu16} with $\rho_1$, $\rho_0$ given to all methods for fair comparison. Additionally, we provide the performance of the ground truth classifier (\emph{true}) trained with uncorrupted labels $(X, y)$, as well as the complete Rank Pruning algorithm (\emph{$\text{RP}_{\rho}$}) trained using the noise rates estimated by Rank Pruning. The top model scores are in bold with $RP_{\rho}$ in red if its performance is better than non-RP models. The $\pi_1=0$ quadrant in each table represents the ``PU learning" case of $\tilde{P}\tilde{N}$ learning. 

\begin{suppfigure}[t]
\begin{center}
\centerline{\includegraphics[width=0.9\columnwidth]{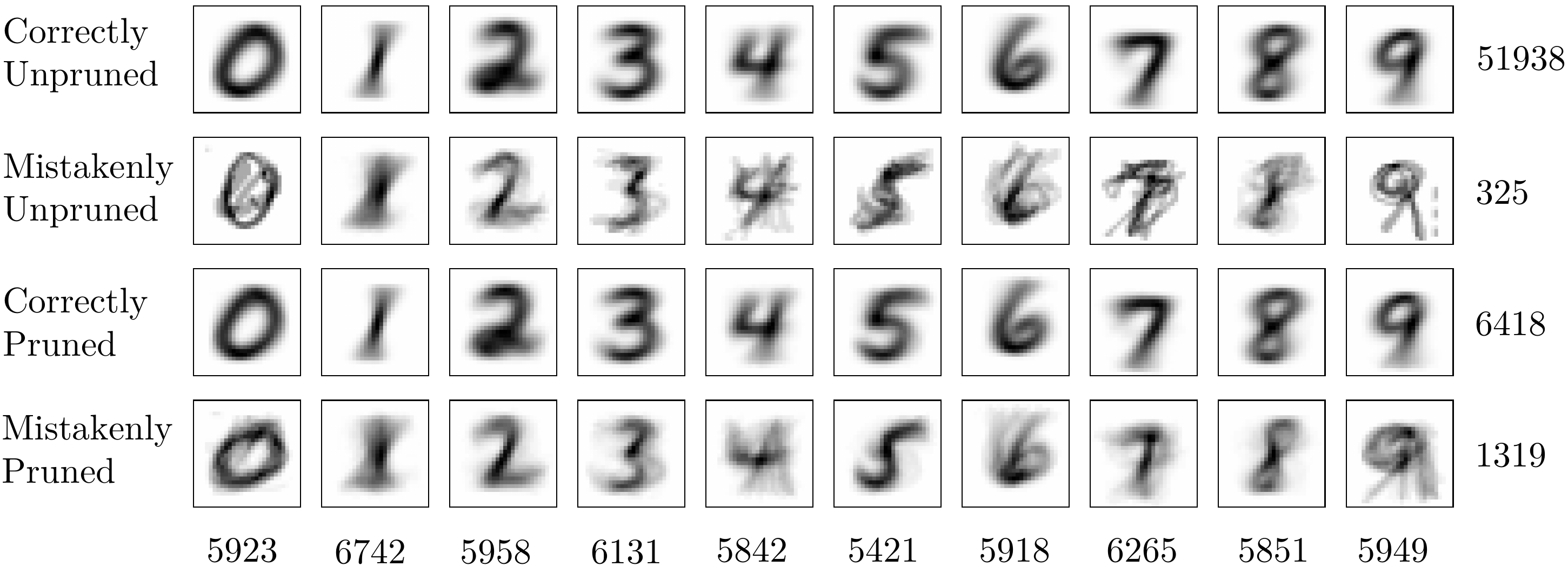}}
\caption{Average image for each digit for the binary classification problem ``1" or ``not 1" in MNIST with logistic regression and significant mislabeling ($\rho_1=0.5, \pi_1=0.5$). The right and bottom numbers count the total number of example images averaged in the corresponding row or column.}
\label{rankprune_mean_mnist}
\end{center}
\end{suppfigure}

Whenever $g(x)=P(\hat{s}=1|x)$ is estimated for any algorithm, we use a 3-fold cross-validation to estimate the probability $g(x)$. For improved performance, a higher fold may be used.

For the logistic regression classifier, we use scikit-learn's LogisticRegression class (\cite{logreg_sklearn}) with default settings (L2 regularization with inverse strength $C=1$).

For the convolutional neural networks (CNN), for MNIST we use the structure in \cite{mnist_cnn_structure} and for CIFAR-10, we use the structure in \cite{cifar_cnn_structure}. A $10\%$ holdout set is used to monitor the weighted validation loss (using the sample weight given by each algorithm) and ends training when there is no decrease for 10 epochs, with a maximum of 50 epochs for MNIST and 150 epochs for CIFAR-10. 

\vskip 0.1in

The following list comprises the MNIST and CIFAR experimental result tables for error, AUC-PR and F1 score metrics:

Table C\ref{table:mnist_logreg_error}: Error for MNIST with logisitic regression as classifier.

Table C\ref{table:mnist_logreg_auc}: AUC-PR for MNIST with logisitic regression as classifier.

Table C\ref{table:mnist_cnn_error}: Error for MNIST with CNN as classifier.

Table C\ref{table:mnist_cnn_auc}: AUC-PR for MNIST with CNN as classifier.

Table C\ref{table:cifar_logreg_f1}: F1 score for CIFAR-10 with logistic regression as classifier.

Table C\ref{table:cifar_logreg_error}: Error for CIFAR-10 with logistic regression as classifier.

Table C\ref{table:cifar_logreg_auc}: AUC-PR for CIFAR-10 with logistic regression as classifier.

Table C\ref{table:cifar_cnn_error}: Error for CIFAR-10 with CNN as classifier.

Table C\ref{table:cifar_cnn_auc}: AUC-PR for CIFAR-10 with CNN as classifier.

\vskip 0.1in
Due to sensitivity to imperfect probability estimation, here \emph{Liu16} always predicts all labels to be positive or negative, resulting in the same metric score for every digit/image in each scenario. Since $p_{y1}\simeq 0.1$, when predicting all labels as positive, \emph{Liu16} has an F1 score of 0.182, error of 0.90, and AUC-PR of 0.55; when predicting all labels as negative, \emph{Liu16} has an F1 score of 0.0, error of 0.1, and AUC-PR of 0.55.

\section{Additional Related Work}

In this section we include tangentially related work which was unable to make it into the final manuscript.

\subsection{One-class classification}

One-class classification \citep{moya_1993_oneclass} is distinguished from binary classification by a training set containing examples from only one class, making it useful for outlier and novelty detection \citep{ Hempstalk:2008:oneclass}. This can be framed as $\tilde{P}\tilde{N}$ learning when outliers take the form of mislabeled examples. The predominant approach, one-class SVM, fits a hyper-boundary around the training class \citep{oneclasssvm1999}, but often performs poorly due to boundary over-sensitivity \citep{Manevitz:2002:OSD:944790.944808} and fails when the training class contains mislabeled examples.

\subsection{\texorpdfstring{$\tilde{P}\tilde{N}$}{PN} learning for Image Recognition and Deep Learning}

Variations of $\tilde{P}\tilde{N}$ learning have been used in the context of machine vision to improve robustness to mislabeling \citep{Xiao2015LearningFM}. In a face recognition task with 90\% of non-faces mislabeled as faces, a bagging model combined with consistency voting was used to remove images with poor voting consistency \citep{angelova2005pruning}. However, no theoretical justification was provided. In the context of deep learning, consistency of predictions for inputs with mislabeling enforces can be enforced by combining a typical cross-entropy loss with an auto-encoder loss \citep{noisy_boostrapping_google}. This method enforces label consistency by constraining the network to uncover the input examples given the output prediction, but is restricted in architecture and generality.

\vskip 0.3in

\begin{supptable*}[th]

\setlength\tabcolsep{2pt} 
\renewcommand{\arraystretch}{0.9}
\caption{Comparison of \textbf{error} for one-vs-rest MNIST (averaged over all digits) using a \textbf{logistic regression} classifier. Except for $RP_\rho$, $\rho_1$, $\rho_0$ are given to all methods. Top model scores are in bold with $RP_\rho$ in red if better (smaller) than non-RP models.} 
\vskip -0.1in
\label{table:mnist_logreg_error}
\begin{center}
\begin{small}
\begin{sc}

\resizebox{\textwidth}{!}{ 
\begin{tabular}{l|rrr|rrrr|rrrr|rrrr}
\toprule

\multicolumn{0}{c}{} & 
\multicolumn{3}{c}{$\pi_1$\textbf{ = 0}}         &   
\multicolumn{4}{c}{$\pi_1$\textbf{ = 0.25}}  & 
\multicolumn{4}{c}{$\pi_1$\textbf{ = 0.5}}  & 
\multicolumn{4}{c}{$\pi_1$\textbf{ = 0.75}}    \\

\textbf{Model,$\rho_1 = $} &      \textbf{0.25} &    \textbf{0.50} &   \textbf{0.75} &   \textbf{0.00} &    \textbf{0.25} &    \textbf{0.50} &   \textbf{0.75} &   \textbf{0.00} &    \textbf{0.25} &    \textbf{0.50} &   \textbf{0.75} &   \textbf{0.00} &    \textbf{0.25} &    \textbf{0.50} &   \textbf{0.75}   \\

\midrule

\textbf{True}   &  0.020 &  0.020 &  0.020 &  0.020 &  0.020 &  0.020 &  0.020 &  0.020 &  0.020 &  0.020 &  0.020 &  0.020 &  0.020 &  0.020 &  0.020 \\
\textbf{RP}$_{\rho}$    &  \textcolor{red}{\textbf{0.023}} &  \textcolor{red}{\textbf{0.025}} &  \textcolor{red}{\textbf{0.031}} &  \textcolor{red}{\textbf{0.024}} &  \textcolor{red}{\textbf{0.025}} &  \textcolor{red}{\textbf{0.027}} &  \textcolor{red}{\textbf{0.038}} &  0.040 &  0.037 &  0.039 &  0.049 &  0.140 &  0.128 &  0.133 &  0.151 \\
\textbf{RP}    &  \textbf{0.022} &  \textbf{0.025} &  \textbf{0.031} &  \textbf{0.021} &  \textbf{0.024} &  \textbf{0.027} &  \textbf{0.035} &  \textbf{0.023} &  \textbf{0.027} &  \textbf{0.031} &  \textbf{0.043} &  \textbf{0.028} &  \textbf{0.036} &  \textbf{0.045} &  0.069 \\
\textbf{Nat13} &  0.025 &  0.030 &  0.038 &  0.025 &  0.029 &  0.034 &  0.042 &  0.030 &  0.033 &  0.038 &  0.047 &  0.035 &  0.039 &  0.046 &  \textbf{0.067} \\
\textbf{Elk08} &  0.025 &  0.030 &  0.038 &  0.026 &  0.028 &  0.032 &  0.042 &  0.030 &  0.031 &  0.035 &  0.051 &  0.092 &  0.093 &  0.123 &  0.189 \\
\textbf{Liu16} &  0.187 &  0.098 &  0.100 &  0.100 &  0.738 &  0.738 &  0.419 &  0.100 &  0.820 &  0.821 &  0.821 &  0.098 &  0.760 &  0.741 &  0.820 \\

\bottomrule
\end{tabular}
}
\end{sc}
\end{small}
\end{center}
\vskip 0.06in
\end{supptable*}

\begin{supptable*}[ht]

\setlength\tabcolsep{2pt} 
\renewcommand{\arraystretch}{0.9}
\caption{Comparison of \textbf{AUC-PR} for one-vs-rest MNIST (averaged over all digits) using a \textbf{logistic regression} classifier. Except for $RP_\rho$, $\rho_1$, $\rho_0$ are given to all methods. Top model scores are in bold with $RP_\rho$ in red if greater than non-RP models.} 
\vskip -0.15in
\label{table:mnist_logreg_auc}
\begin{center}
\begin{small}
\begin{sc}

\resizebox{\textwidth}{!}{ 
\begin{tabular}{l|rrr|rrrr|rrrr|rrrr}
\toprule

\multicolumn{0}{c}{} & 
\multicolumn{3}{c}{$\pi_1$\textbf{ = 0}}         &   
\multicolumn{4}{c}{$\pi_1$\textbf{ = 0.25}}  & 
\multicolumn{4}{c}{$\pi_1$\textbf{ = 0.5}}  & 
\multicolumn{4}{c}{$\pi_1$\textbf{ = 0.75}}    \\

\textbf{Model,$\rho_1 = $} &      \textbf{0.25} &    \textbf{0.50} &   \textbf{0.75} &   \textbf{0.00} &    \textbf{0.25} &    \textbf{0.50} &   \textbf{0.75} &   \textbf{0.00} &    \textbf{0.25} &    \textbf{0.50} &   \textbf{0.75} &   \textbf{0.00} &    \textbf{0.25} &    \textbf{0.50} &   \textbf{0.75}   \\

\midrule

\textbf{True}   &  0.935 &  0.935 &  0.935 &  0.935 &  0.935 &  0.935 &  0.935 &  0.935 &  0.935 &  0.935 &  0.935 &  0.935 &  0.935 &  0.935 &  0.935 \\
\textbf{RP}$_{\rho}$ &  0.921 &  \textcolor{red}{\textbf{0.913}} &  \textcolor{red}{\textbf{0.882}} &  \textcolor{red}{\textbf{0.928}} &  \textcolor{red}{\textbf{0.920}} &  \textcolor{red}{\textbf{0.906}} &  \textcolor{red}{\textbf{0.853}} &  \textcolor{red}{\textbf{0.903}} &  \textcolor{red}{\textbf{0.902}} &  \textcolor{red}{\textbf{0.879}} &  \textcolor{red}{\textbf{0.803}} &  0.851 &  0.835 &  \textcolor{red}{\textbf{0.788}} &  0.640 \\
\textbf{RP}    &  \textbf{0.922} &  \textbf{0.913} &  \textbf{0.882} &  \textbf{0.930} &  \textbf{0.921} &  \textbf{0.906} &  \textbf{0.858} &  \textbf{0.922} &  \textbf{0.903} &  \textbf{0.883} &  \textbf{0.811} &  \textbf{0.893} &  \textbf{0.841} &  \textbf{0.799} &  0.621 \\
\textbf{Nat13} &  \textbf{0.922} &  0.908 &  0.878 &  0.918 &  0.909 &  0.890 &  0.839 &  0.899 &  0.892 &  0.862 &  0.794 &  0.863 &  0.837 &  0.784 &  \textbf{0.645} \\
\textbf{Elk08} &  0.921 &  0.903 &  0.864 &  0.917 &  0.908 &  0.884 &  0.821 &  0.898 &  0.892 &  0.861 &  0.763 &  0.852 &  0.837 &  0.772 &  0.579 \\
\textbf{Liu16} &  0.498 &  0.549 &  0.550 &  0.550 &  0.500 &  0.550 &  0.505 &  0.550 &  0.550 &  0.550 &  0.549 &  0.503 &  0.512 &  0.550 &  0.550 \\

\bottomrule
\end{tabular}
}
\end{sc}
\end{small}
\end{center}
\vskip 0.08in
\end{supptable*}


\begin{supptable*}[ht]

\setlength\tabcolsep{1pt} 
\renewcommand{\arraystretch}{0.85}
\caption{Comparison of \textbf{error} for one-vs-rest MNIST (averaged over all digits) using a \textbf{CNN} classifier. Except for $RP_\rho$, $\rho_1$, $\rho_0$ are given to all methods. Top model scores are in bold with $RP_\rho$ in red if better (smaller) than non-RP models.} 
\vskip -0.1in
\label{table:mnist_cnn_error}
\begin{center}
\begin{small}
\begin{sc}

\resizebox{\textwidth}{!}{
\begin{tabular}{l|c|ccccc|ccccc|ccccc|ccccc}
\toprule

\multicolumn{1}{c}{} &  
\multicolumn{1}{c|}{} & 
\multicolumn{5}{c|}{$\pi_1$\textbf{ = 0}}   &   
\multicolumn{5}{c|}{$\pi_1$\textbf{ = 0.25}}  & 
\multicolumn{10}{c}{$\pi_1$\textbf{ = 0.5}}   \\

\multicolumn{1}{c}{} &  
\multicolumn{1}{c|}{} & 
\multicolumn{5}{c|}{$\rho_1$\textbf{ = 0.5}} & 
\multicolumn{5}{c|}{$\rho_1$\textbf{ = 0.25}}  & 
\multicolumn{5}{c}{$\rho_1$\textbf{ = 0}} & 
\multicolumn{5}{c}{$\rho_1$\textbf{ = 0.5}}    \\

{\textbf{IMAGE}} &   \textbf{True} & \textbf{RP}$_{\rho}$ &    \textbf{RP} & \textbf{Nat13} & \textbf{Elk08} & \textbf{Liu16} & \textbf{RP}$_{\rho}$ &    \textbf{RP} & \textbf{Nat13} & \textbf{Elk08} & \textbf{Liu16} & \textbf{RP}$_{\rho}$ &    \textbf{RP} & \textbf{Nat13} & \textbf{Elk08} & \textbf{Liu16} & \textbf{RP}$_{\rho}$ &    \textbf{RP} & \textbf{Nat13} & \textbf{Elk08} & \textbf{Liu16} \\
\midrule

\textbf{0}     &  0.0013 &  \textcolor{red}{\textbf{0.0018}} &  \textbf{0.0023} &  0.0045 &  0.0047 &  0.9020 &  \textcolor{red}{\textbf{0.0017}} &  \textbf{0.0016} &  0.0034 &  0.0036 &  0.9020 &  \textcolor{red}{\textbf{0.0017}} &  \textbf{0.0016} &  0.0031 &  0.0026 &  0.0029 &  \textcolor{red}{\textbf{0.0021}} &  \textbf{0.0022} &  0.0116 &  0.0069 &  0.9020 \\
\textbf{1}     &  0.0015 &  \textcolor{red}{\textbf{0.0022}} &  \textbf{0.0020} &  0.0025 &  0.0034 &  0.8865 &  \textcolor{red}{\textbf{0.0019}} &  \textbf{0.0019} &  0.0035 &  0.0030 &  0.8865 &  0.0023 &  0.0020 &  0.0018 &  \textbf{0.0016} &  0.0023 &  \textcolor{red}{\textbf{0.0025}} &  \textbf{0.0025} &  0.0036 &  0.0027 &  0.8865 \\
\textbf{2}     &  0.0027 &  \textcolor{red}{\textbf{0.0054}} &  \textbf{0.0049} &  0.0057 &  0.0062 &  0.8968 &  \textcolor{red}{\textbf{0.0032}} &  \textbf{0.0035} &  0.0045 &  0.0051 &  0.8968 &  0.0030 &  0.0029 &  0.0031 &  0.0029 &  \textbf{0.0024} &  \textcolor{red}{\textbf{0.0059}} &  \textbf{0.0050} &  0.0066 &  0.0083 &  0.8968 \\
\textbf{3}     &  0.0020 &  \textcolor{red}{\textbf{0.0032}} &  \textbf{0.0032} &  0.0055 &  0.0038 &  0.8990 &  \textcolor{red}{\textbf{0.0029}} &  \textbf{0.0029} &  0.0043 &  0.0043 &  0.8990 &  \textcolor{red}{\textbf{0.0021}} &  0.0027 &  \textbf{0.0023} &  \textbf{0.0023} &  0.0032 &  \textcolor{red}{\textbf{0.0038}} &  \textbf{0.0042} &  0.0084 &  0.0057 &  0.8990 \\
\textbf{4}     &  0.0012 &  \textcolor{red}{\textbf{0.0037}} &  0.0040 &  \textbf{0.0038} &  0.0044 &  0.9018 &  \textcolor{red}{\textbf{0.0029}} &  \textbf{0.0025} &  0.0055 &  0.0069 &  0.9018 &  0.0026 &  0.0020 &  \textbf{0.0019} &  0.0021 &  0.0030 &  \textcolor{red}{\textbf{0.0044}} &  \textbf{0.0035} &  0.0086 &  0.0077 &  0.9018 \\
\textbf{5}     &  0.0019 &  \textcolor{red}{\textbf{0.0032}} &  \textbf{0.0035} &  0.0039 &  0.0038 &  0.9108 &  \textcolor{red}{\textbf{0.0027}} &  \textbf{0.0031} &  0.0062 &  0.0060 &  0.9108 &  \textcolor{red}{\textbf{0.0021}} &  0.0024 &  0.0024 &  0.0028 &  \textbf{0.0023} &  \textcolor{red}{\textbf{0.0061}} &  \textbf{0.0056} &  0.0066 &  0.0074 &  0.9108 \\
\textbf{6}     &  0.0021 &  \textcolor{red}{\textbf{0.0027}} &  \textbf{0.0028} &  0.0053 &  0.0035 &  0.9042 &  \textcolor{red}{\textbf{0.0028}} &  \textbf{0.0025} &  0.0042 &  0.0036 &  0.9042 &  0.0029 &  0.0029 &  \textbf{0.0022} &  0.0024 &  0.0028 &  \textcolor{red}{\textbf{0.0032}} &  \textbf{0.0035} &  0.0098 &  0.0075 &  0.9042 \\
\textbf{7}     &  0.0026 &  \textcolor{red}{\textbf{0.0039}} &  \textbf{0.0041} &  0.0066 &  0.0103 &  0.8972 &  \textcolor{red}{\textbf{0.0050}} &  \textbf{0.0052} &  0.0058 &  0.0058 &  0.8972 &  0.0049 &  0.0040 &  \textbf{0.0030} &  0.0037 &  0.0035 &  \textcolor{red}{\textbf{0.0054}} &  \textbf{0.0064} &  0.0113 &  0.0085 &  0.8972 \\
\textbf{8}     &  0.0022 &  \textcolor{red}{\textbf{0.0047}} &  \textbf{0.0043} &  0.0106 &  0.0063 &  0.9026 &  \textcolor{red}{\textbf{0.0034}} &  \textbf{0.0036} &  0.0062 &  0.0091 &  0.9026 &  0.0036 &  \textbf{0.0030} &  0.0035 &  0.0041 &  0.0032 &  \textcolor{red}{\textbf{0.0044}} &  \textbf{0.0048} &  0.0234 &  0.0077 &  0.9026 \\
\textbf{9}     &  0.0036 &  0.0067 &  \textbf{0.0052} &  0.0056 &  0.0124 &  0.8991 &  \textcolor{red}{\textbf{0.0048}} &  \textbf{0.0051} &  0.0065 &  0.0064 &  0.8991 &  0.0048 &  0.0050 &  0.0051 &  \textbf{0.0043} &  0.0059 &  \textcolor{red}{\textbf{0.0081}} &  0.0114 &  0.0131 &  \textbf{0.0112} &  0.8991 \\
\midrule
\textbf{AVG} &  0.0021 &  \textcolor{red}{\textbf{0.0038}} &  \textbf{0.0036} &  0.0054 &  0.0059 &  0.9000 &  \textcolor{red}{\textbf{0.0031}} &  \textbf{0.0032} &  0.0050 &  0.0054 &  0.9000 &  0.0030 &  \textbf{0.0028} &  \textbf{0.0028} &  0.0029 &  0.0032 &  \textcolor{red}{\textbf{0.0046}} &  \textbf{0.0049} &  0.0103 &  0.0074 &  0.9000 \\

\bottomrule
\end{tabular}
}
\end{sc}
\end{small}
\end{center}
\end{supptable*}

\begin{supptable*}[ht]

\setlength\tabcolsep{1pt} 
\renewcommand{\arraystretch}{0.85}
\caption{Comparison of \textbf{AUC-PR} for one-vs-rest MNIST (averaged over all digits) using a \textbf{CNN} classifier. Except for $RP_\rho$, $\rho_1$, $\rho_0$ are given to all methods. Top model scores are in bold with $RP_\rho$ in red if greater than non-RP models.} 
\vskip -0.1in
\label{table:mnist_cnn_auc}
\begin{center}
\begin{small}
\begin{sc}

\resizebox{\textwidth}{!}{
\begin{tabular}{l|c|ccccc|ccccc|ccccc|ccccc}
\toprule

\multicolumn{1}{c}{} &  
\multicolumn{1}{c|}{} & 
\multicolumn{5}{c|}{$\pi_1$\textbf{ = 0}}   &   
\multicolumn{5}{c|}{$\pi_1$\textbf{ = 0.25}}  & 
\multicolumn{10}{c}{$\pi_1$\textbf{ = 0.5}}   \\

\multicolumn{1}{c}{} &  
\multicolumn{1}{c|}{} & 
\multicolumn{5}{c|}{$\rho_1$\textbf{ = 0.5}} & 
\multicolumn{5}{c|}{$\rho_1$\textbf{ = 0.25}}  & 
\multicolumn{5}{c}{$\rho_1$\textbf{ = 0}} & 
\multicolumn{5}{c}{$\rho_1$\textbf{ = 0.5}}    \\

{\textbf{IMAGE}} &   \textbf{True} & \textbf{RP}$_{\rho}$ &    \textbf{RP} & \textbf{Nat13} & \textbf{Elk08} & \textbf{Liu16} & \textbf{RP}$_{\rho}$ &    \textbf{RP} & \textbf{Nat13} & \textbf{Elk08} & \textbf{Liu16} & \textbf{RP}$_{\rho}$ &    \textbf{RP} & \textbf{Nat13} & \textbf{Elk08} & \textbf{Liu16} & \textbf{RP}$_{\rho}$ &    \textbf{RP} & \textbf{Nat13} & \textbf{Elk08} & \textbf{Liu16} \\
\midrule
\textbf{0}     & 0.9998 & \textcolor{red}{\textbf{0.9992}} & \textbf{0.9990} & 0.9986 & 0.9982 & 0.5490 & \textcolor{red}{\textbf{0.9996}} & \textbf{0.9996} & 0.9986 & 0.9979 & 0.5490 & \textcolor{red}{\textbf{0.9989}} & \textbf{0.9995} & 0.9976 & 0.9979 & 0.9956 & \textcolor{red}{\textbf{0.9984}} & \textbf{0.9982} & 0.9963 & 0.9928 & 0.5490 \\
\textbf{1}     & 0.9999 & \textcolor{red}{\textbf{0.9995}} & \textbf{0.9995} & 0.9976 & 0.9974 & 0.5568 & \textcolor{red}{\textbf{0.9996}} & 0.9993 & \textbf{0.9995} & \textbf{0.9995} & 0.5568 & \textcolor{red}{\textbf{0.9995}} & \textbf{0.9998} & 0.9982 & 0.9972 & 0.9965 & \textcolor{red}{\textbf{0.9995}} & \textbf{0.9994} & 0.9978 & 0.9985 & 0.5568 \\
\textbf{2}     & 0.9994 & \textcolor{red}{\textbf{0.9971}} & \textbf{0.9969} & 0.9917 & 0.9942 & 0.5516 & \textcolor{red}{\textbf{0.9980}} & \textbf{0.9977} & 0.9971 & 0.9945 & 0.5516 & \textcolor{red}{\textbf{0.9988}} & \textbf{0.9992} & 0.9958 & 0.9934 & 0.9940 & \textcolor{red}{\textbf{0.9938}} & \textbf{0.9947} & 0.9847 & 0.9873 & 0.5516 \\
\textbf{3}     & 0.9996 & \textcolor{red}{\textbf{0.9986}} & \textbf{0.9987} & 0.9983 & 0.9984 & 0.5505 & \textcolor{red}{\textbf{0.9991}} & \textbf{0.9989} & 0.9982 & 0.9980 & 0.5505 & \textcolor{red}{\textbf{0.9993}} & \textbf{0.9994} & 0.9991 & 0.9971 & 0.9974 & \textcolor{red}{\textbf{0.9969}} & \textbf{0.9959} & 0.9951 & \textbf{0.9959} & 0.5505 \\
\textbf{4}     & 0.9997 & 0.9982 & \textbf{0.9989} & 0.9939 & 0.9988 & 0.0891 & \textcolor{red}{\textbf{0.9992}} & \textbf{0.9991} & 0.9976 & 0.9965 & 0.5491 & \textcolor{red}{\textbf{0.9994}} & \textbf{0.9996} & 0.9985 & 0.9978 & 0.9986 & \textcolor{red}{\textbf{0.9983}} & \textbf{0.9977} & 0.9961 & 0.9919 & 0.5491 \\
\textbf{5}     & 0.9993 & \textcolor{red}{\textbf{0.9982}} & \textbf{0.9976} & 0.9969 & 0.9956 & 0.5446 & \textcolor{red}{\textbf{0.9986}} & \textbf{0.9987} & 0.9983 & 0.9979 & 0.5446 & \textcolor{red}{\textbf{0.9984}} & \textbf{0.9982} & 0.9971 & 0.9963 & 0.9929 & \textcolor{red}{\textbf{0.9958}} & \textbf{0.9965} & 0.9946 & 0.9934 & 0.5446 \\
\textbf{6}     & 0.9987 & \textcolor{red}{\textbf{0.9976}} & \textbf{0.9970} & 0.9928 & 0.9931 & 0.5479 & \textcolor{red}{\textbf{0.9974}} & \textbf{0.9980} & 0.9956 & 0.9959 & 0.5479 & \textcolor{red}{\textbf{0.9968}} & \textbf{0.9983} & 0.9933 & 0.9950 & 0.9905 & \textcolor{red}{\textbf{0.9964}} & 0.9957 & 0.9942 & \textbf{0.9961} & 0.5479 \\
\textbf{7}     & 0.9989 & \textcolor{red}{\textbf{0.9973}} & \textbf{0.9972} & 0.9965 & 0.9944 & 0.0721 & 0.9968 & 0.9973 & 0.9966 & \textbf{0.9979} & 0.5514 & 0.9969 & \textbf{0.9983} & 0.9961 & 0.9958 & 0.9974 & \textcolor{red}{\textbf{0.9933}} & \textbf{0.9937} & 0.9896 & 0.9886 & 0.5514 \\
\textbf{8}     & 0.9996 & \textcolor{red}{\textbf{0.9974}} & \textbf{0.9964} & \textbf{0.9964} & 0.9946 & 0.5487 & \textcolor{red}{\textbf{0.9981}} & \textbf{0.9981} & 0.9973 & 0.9971 & 0.5487 & \textbf{0.9983} & 0.9988 & 0.9984 & 0.9976 & \textbf{0.9989} & \textcolor{red}{\textbf{0.9976}} & \textbf{0.9975} & 0.9873 & 0.9893 & 0.5487 \\
\textbf{9}     & 0.9979 & \textcolor{red}{\textbf{0.9931}} & \textbf{0.9951} & 0.9901 & 0.9922 & 0.5504 & \textcolor{red}{\textbf{0.9935}} & \textbf{0.9951} & 0.9933 & 0.9920 & 0.5504 & \textcolor{red}{\textbf{0.9961}} & \textbf{0.9951} & 0.9924 & 0.9922 & 0.9912 & \textcolor{red}{\textbf{0.9877}} & \textbf{0.9876} & 0.9819 & 0.9828 & 0.5504 \\
\midrule
\textbf{AVG} & 0.9993 & \textcolor{red}{\textbf{0.9976}} & \textbf{0.9976} & 0.9953 & 0.9957 & 0.4561 & \textcolor{red}{\textbf{0.9980}} & \textbf{0.9982} & 0.9972 & 0.9967 & 0.5500 & \textcolor{red}{\textbf{0.9983}} & \textbf{0.9986} & 0.9966 & 0.9960 & 0.9953 & \textcolor{red}{\textbf{0.9958}} & \textbf{0.9957} & 0.9918 & 0.9917 & 0.5500 \\

\bottomrule
\end{tabular}
}
\end{sc}
\end{small}
\end{center}
\vskip -0.08in
\end{supptable*}

\begin{supptable*}[ht]

\setlength\tabcolsep{1pt} 
\renewcommand{\arraystretch}{0.85}
\caption{Comparison of \textbf{F1 score} for one-vs-rest CIFAR-10 (averaged over all images) using a \textbf{logistic regression} classifier. Except for $RP_\rho$, $\rho_1$, $\rho_0$ are given to all methods. Top model scores are in bold with $RP_\rho$ in red if greater than non-RP models.} 
\vskip -0.1in
\label{table:cifar_logreg_f1}
\begin{center}
\begin{small}
\begin{sc}

\resizebox{\textwidth}{!}{
\begin{tabular}{l|c|ccccc|ccccc|ccccc|ccccc}
\toprule

\multicolumn{1}{c}{} &  
\multicolumn{1}{c|}{} & 
\multicolumn{5}{c|}{$\pi_1$\textbf{ = 0}}   &   
\multicolumn{5}{c|}{$\pi_1$\textbf{ = 0.25}}  & 
\multicolumn{10}{c}{$\pi_1$\textbf{ = 0.5}}   \\

\multicolumn{1}{c}{} &  
\multicolumn{1}{c|}{} & 
\multicolumn{5}{c|}{$\rho_1$\textbf{ = 0.5}} & 
\multicolumn{5}{c|}{$\rho_1$\textbf{ = 0.25}}  & 
\multicolumn{5}{c}{$\rho_1$\textbf{ = 0}} & 
\multicolumn{5}{c}{$\rho_1$\textbf{ = 0.5}}    \\

{\textbf{IMAGE}} &   \textbf{True} & \textbf{RP}$_{\rho}$ &    \textbf{RP} & \textbf{Nat13} & \textbf{Elk08} & \textbf{Liu16} & \textbf{RP}$_{\rho}$ &    \textbf{RP} & \textbf{Nat13} & \textbf{Elk08} & \textbf{Liu16} & \textbf{RP}$_{\rho}$ &    \textbf{RP} & \textbf{Nat13} & \textbf{Elk08} & \textbf{Liu16} & \textbf{RP}$_{\rho}$ &    \textbf{RP} & \textbf{Nat13} & \textbf{Elk08} & \textbf{Liu16} \\
\midrule
\textbf{plane}   &  0.272 &  \textcolor{red}{\textbf{0.311}} &  \textbf{0.252} &  0.217 &  0.220 &  0.182 &  \textcolor{red}{\textbf{0.329}} &  \textbf{0.275} &  0.222 &  0.224 &  0.182 &  \textcolor{red}{\textbf{0.330}} &  \textbf{0.265} &  0.231 &  0.259 &    0.0 &  \textcolor{red}{\textbf{0.266}} &  \textbf{0.188} &  0.183 &  0.187 &  0.182 \\
\textbf{auto} &  0.374 &  \textcolor{red}{\textbf{0.389}} &  \textbf{0.355} &  0.318 &  0.320 &  0.182 &  \textcolor{red}{\textbf{0.388}} &  \textbf{0.368} &  0.321 &  0.328 &  0.182 &  \textcolor{red}{\textbf{0.372}} &  \textbf{0.355} &  0.308 &  0.341 &    0.0 &  \textcolor{red}{\textbf{0.307}} &  0.287 &  0.287 &  \textbf{0.297} &  0.182 \\
\textbf{bird}       &  0.136 &  \textcolor{red}{\textbf{0.241}} &  0.167 &  0.143 &  0.136 &  \textbf{0.182} &  \textcolor{red}{\textbf{0.248}} &  \textbf{0.185} &  0.137 &  0.137 &  0.182 &  \textcolor{red}{\textbf{0.258}} &  \textbf{0.147} &  0.100 &  0.126 &    0.0 &  \textcolor{red}{\textbf{0.206}} &  0.153 &  0.132 &  0.150 &  \textbf{0.182} \\
\textbf{cat}        &  0.122 &  \textcolor{red}{\textbf{0.246}} &  0.170 &  0.141 &  0.150 &  \textbf{0.182} &  \textcolor{red}{\textbf{0.232}} &  0.163 &  0.112 &  0.127 &  \textbf{0.182} &  \textcolor{red}{\textbf{0.241}} &  \textbf{0.125} &  0.068 &  0.103 &    0.0 &  \textcolor{red}{\textbf{0.209}} &  0.148 &  0.119 &  0.157 &  \textbf{0.182} \\
\textbf{deer}       &  0.166 &  \textcolor{red}{\textbf{0.250}} &  \textbf{0.184} &  0.153 &  0.164 &  0.182 &  \textcolor{red}{\textbf{0.259}} &  0.175 &  0.146 &  0.163 &  \textbf{0.182} &  \textcolor{red}{\textbf{0.259}} &  \textbf{0.177} &  0.126 &  0.164 &    0.0 &  \textcolor{red}{\textbf{0.222}} &  0.162 &  0.132 &  0.164 &  \textbf{0.182} \\
\textbf{dog}        &  0.139 &  \textcolor{red}{\textbf{0.245}} &  0.174 &  0.146 &  0.148 &  \textbf{0.182} &  \textcolor{red}{\textbf{0.262}} &  0.171 &  0.115 &  0.126 &  \textbf{0.182} &  \textcolor{red}{\textbf{0.254}} &  \textbf{0.152} &  0.075 &  0.120 &    0.0 &  \textcolor{red}{\textbf{0.203}} &  0.151 &  0.128 &  0.137 &  \textbf{0.182} \\
\textbf{frog}       &  0.317 &  \textcolor{red}{\textbf{0.322}} &  \textbf{0.315} &  0.289 &  0.281 &  0.182 &  \textcolor{red}{\textbf{0.350}} &  \textbf{0.319} &  0.283 &  0.299 &  0.182 &  \textcolor{red}{\textbf{0.346}} &  \textbf{0.305} &  0.239 &  0.279 &    0.0 &  \textcolor{red}{\textbf{0.308}} &  0.252 &  0.244 &  \textbf{0.269} &  0.182 \\
\textbf{horse}      &  0.300 &  \textcolor{red}{\textbf{0.300}} &  \textbf{0.299} &  0.283 &  0.263 &  0.182 &  \textcolor{red}{\textbf{0.334}} &  \textbf{0.313} &  0.272 &  0.281 &  0.182 &  \textcolor{red}{\textbf{0.322}} &  \textbf{0.310} &  0.260 &  0.292 &    0.0 &  \textcolor{red}{\textbf{0.275}} &  \textbf{0.258} &  0.240 &  0.245 &  0.182 \\
\textbf{ship}       &  0.322 &  \textcolor{red}{\textbf{0.343}} &  \textbf{0.322} &  0.297 &  0.272 &  0.182 &  \textcolor{red}{\textbf{0.385}} &  \textbf{0.319} &  0.287 &  0.289 &  0.182 &  \textcolor{red}{\textbf{0.350}} &  \textbf{0.303} &  0.250 &  0.293 &    0.0 &  \textcolor{red}{\textbf{0.304}} &  \textbf{0.248} &  0.230 &  0.237 &  0.182 \\
\textbf{truck}      &  0.330 &  \textcolor{red}{\textbf{0.359}} &  \textbf{0.323} &  0.273 &  0.261 &  0.182 &  \textcolor{red}{\textbf{0.369}} &  \textbf{0.327} &  0.293 &  0.290 &  0.182 &  \textcolor{red}{\textbf{0.343}} &  \textbf{0.302} &  0.278 &  0.299 &    0.0 &  \textcolor{red}{\textbf{0.313}} &  0.246 &  0.252 &  \textbf{0.262} &  0.182 \\
\midrule
\textbf{AVG}      &  0.248 &  \textcolor{red}{\textbf{0.301}} &  \textbf{0.256} &  0.226 &  0.221 &  0.182 &  \textcolor{red}{\textbf{0.316}} &  \textbf{0.262} &  0.219 &  0.226 &  0.182 &  \textcolor{red}{\textbf{0.308}} &  \textbf{0.244} &  0.194 &  0.228 &   0.000 &  \textcolor{red}{\textbf{0.261}} &  0.209 &  0.195 &  \textbf{0.210} &  0.182 \\
\bottomrule
\end{tabular}
}
\end{sc}
\end{small}
\end{center}
\vskip -0.1in
\end{supptable*}

\begin{supptable*}[ht]

\setlength\tabcolsep{1pt} 
\renewcommand{\arraystretch}{0.85}
\caption{Comparison of \textbf{error} for one-vs-rest CIFAR-10 (averaged over all images) using a \textbf{logistic regression} classifier. Except for $RP_\rho$, $\rho_1$, $\rho_0$ are given to all methods. Top model scores are in bold with $RP_\rho$ in red if better (smaller) than non-RP models. Here the logistic regression classifier severely underfits CIFAR, resulting in Rank Pruning pruning out some correctly labeled examples that ``confuse" the classifier, hence in this scenario, RP and RP$_{\rho}$ generally have slightly smaller precision, much higher recall, and hence larger F1 scores than other models and even the ground truth classifier (Table C\ref{table:cifar_logreg_f1}). Due to the class inbalance ($p_{y1}=0.1$) and their larger recall, RP and RP$_{\rho}$ here have larger error than the other models.} 
\vskip -0.1in
\label{table:cifar_logreg_error}
\begin{center}
\begin{small}
\begin{sc}

\resizebox{\textwidth}{!}{
\begin{tabular}{l|c|ccccc|ccccc|ccccc|ccccc}
\toprule

\multicolumn{1}{c}{} &  
\multicolumn{1}{c|}{} & 
\multicolumn{5}{c|}{$\pi_1$\textbf{ = 0}}   &   
\multicolumn{5}{c|}{$\pi_1$\textbf{ = 0.25}}  & 
\multicolumn{10}{c}{$\pi_1$\textbf{ = 0.5}}   \\

\multicolumn{1}{c}{} &  
\multicolumn{1}{c|}{} & 
\multicolumn{5}{c|}{$\rho_1$\textbf{ = 0.5}} & 
\multicolumn{5}{c|}{$\rho_1$\textbf{ = 0.25}}  & 
\multicolumn{5}{c}{$\rho_1$\textbf{ = 0}} & 
\multicolumn{5}{c}{$\rho_1$\textbf{ = 0.5}}    \\

{\textbf{IMAGE}} &   \textbf{True} & \textbf{RP}$_{\rho}$ &    \textbf{RP} & \textbf{Nat13} & \textbf{Elk08} & \textbf{Liu16} & \textbf{RP}$_{\rho}$ &    \textbf{RP} & \textbf{Nat13} & \textbf{Elk08} & \textbf{Liu16} & \textbf{RP}$_{\rho}$ &    \textbf{RP} & \textbf{Nat13} & \textbf{Elk08} & \textbf{Liu16} & \textbf{RP}$_{\rho}$ &    \textbf{RP} & \textbf{Nat13} & \textbf{Elk08} & \textbf{Liu16} \\
\midrule
\textbf{plane}   & 0.107 &  0.287 & 0.133 &  0.123 &  \textbf{0.122} &  0.900 &  0.177 & 0.128 &  \textbf{0.119} &  0.123 &  0.900 &  0.248 & 0.124 &  0.110 &  0.118 &  \textbf{0.100} &  0.202 & 0.147 &  \textbf{0.142} &  0.160 &  0.900 \\
\textbf{auto} & 0.099 &  0.184 & 0.120 &  \textbf{0.110} &  \textbf{0.110} &  0.900 &  0.132 & 0.114 &  \textbf{0.105} &  0.109 &  0.900 &  0.189 & 0.110 &  0.105 &  0.110 &  \textbf{0.100} &  0.159 & 0.129 &  \textbf{0.125} &  0.139 &  0.900 \\
\textbf{bird}       & 0.117 &  0.354 & 0.148 &  0.133 &  \textbf{0.131} &  0.900 &  0.217 & 0.135 &  \textbf{0.120} &  0.125 &  0.900 &  0.277 & 0.135 &  0.115 &  0.123 &  \textbf{0.100} &  0.226 & 0.147 &  \textbf{0.139} &  0.158 &  0.900 \\
\textbf{cat}        & 0.114 &  0.351 & 0.138 &  \textbf{0.129} &  \textbf{0.129} &  0.900 &  0.208 & 0.139 &  \textbf{0.122} &  0.125 &  0.900 &  0.303 & 0.132 &  0.114 &  0.122 &  \textbf{0.100} &  0.225 & 0.151 &  \textbf{0.141} &  0.158 &  0.900 \\
\textbf{deer}       & 0.112 &  0.336 & 0.143 &  \textbf{0.128} &  0.130 &  0.900 &  0.194 & 0.135 &  \textbf{0.120} &  0.122 &  0.900 &  0.271 & 0.133 &  0.118 &  0.126 &  \textbf{0.100} &  0.209 & 0.150 &  \textbf{0.147} &  0.161 &  0.900 \\
\textbf{dog}        & 0.119 &  0.370 & 0.150 &  \textbf{0.136} &  0.138 &  0.900 &  0.205 & 0.142 &  \textbf{0.129} &  0.132 &  0.900 &  0.288 & 0.135 &  0.120 &  0.128 &  \textbf{0.100} &  0.229 & 0.154 &  \textbf{0.147} &  0.168 &  0.900 \\
\textbf{frog}       & 0.107 &  0.228 & 0.128 &  \textbf{0.117} &  \textbf{0.117} &  0.900 &  0.155 & 0.124 &  \textbf{0.113} &  0.115 &  0.900 &  0.228 & 0.118 &  0.110 &  0.116 &  \textbf{0.100} &  0.167 & 0.137 &  \textbf{0.130} &  0.142 &  0.900 \\
\textbf{horse}      & 0.104 &  0.251 & 0.127 &  \textbf{0.114} &  0.116 &  0.900 &  0.153 & 0.123 &  \textbf{0.110} &  0.112 &  0.900 &  0.224 & 0.116 &  0.108 &  0.113 &  \textbf{0.100} &  0.178 & 0.134 &  \textbf{0.129} &  0.144 &  0.900 \\
\textbf{ship}       & 0.112 &  0.239 & 0.134 &  \textbf{0.121} &  0.126 &  0.900 &  0.160 & 0.131 &  \textbf{0.119} &  0.123 &  0.900 &  0.236 & 0.122 &  0.113 &  0.120 &  \textbf{0.100} &  0.193 & 0.145 &  \textbf{0.139} &  0.159 &  0.900 \\
\textbf{truck}      & 0.106 &  0.210 & 0.130 &  \textbf{0.121} &  0.122 &  0.900 &  0.145 & 0.125 &  \textbf{0.113} &  0.117 &  0.900 &  0.213 & 0.121 &  0.108 &  0.117 &  \textbf{0.100} &  0.165 & 0.142 &  \textbf{0.134} &  0.150 &  0.900 \\
\midrule
\textbf{AVG}      & 0.110 &  0.281 & 0.135 &  \textbf{0.123} &  0.124 &  0.900 &  0.175 & 0.130 &  \textbf{0.117} &  0.120 &  0.900 &  0.248 & 0.125 &  0.112 &  0.119 &  \textbf{0.100} &  0.195 & 0.144 &  \textbf{0.137} &  0.154 &  0.900 \\

\bottomrule
\end{tabular}
}
\end{sc}
\end{small}
\end{center}
\end{supptable*}

\begin{supptable*}[ht]

\setlength\tabcolsep{1pt} 
\renewcommand{\arraystretch}{0.85}
\caption{Comparison of \textbf{AUC-PR} for one-vs-rest CIFAR-10 (averaged over all images) using a \textbf{logistic regression} classifier. Except for $RP_\rho$, $\rho_1$, $\rho_0$ are given to all methods. Top model scores are in bold with $RP_\rho$ in red if greater than non-RP models. Since $p_{y1}=0.1$, here \emph{Liu16} always predicts all labels as positive or negative, resulting in a constant AUC-PR of 0.550.} 
\vskip -0.1in
\label{table:cifar_logreg_auc}
\begin{center}
\begin{small}
\begin{sc}

\resizebox{\textwidth}{!}{
\begin{tabular}{l|c|ccccc|ccccc|ccccc|ccccc}
\toprule

\multicolumn{1}{c}{} &  
\multicolumn{1}{c|}{} & 
\multicolumn{5}{c|}{$\pi_1$\textbf{ = 0}}   &   
\multicolumn{5}{c|}{$\pi_1$\textbf{ = 0.25}}  & 
\multicolumn{10}{c}{$\pi_1$\textbf{ = 0.5}}   \\

\multicolumn{1}{c}{} &  
\multicolumn{1}{c|}{} & 
\multicolumn{5}{c|}{$\rho_1$\textbf{ = 0.5}} & 
\multicolumn{5}{c|}{$\rho_1$\textbf{ = 0.25}}  & 
\multicolumn{5}{c}{$\rho_1$\textbf{ = 0}} & 
\multicolumn{5}{c}{$\rho_1$\textbf{ = 0.5}}    \\

{\textbf{IMAGE}} &   \textbf{True} & \textbf{RP}$_{\rho}$ &    \textbf{RP} & \textbf{Nat13} & \textbf{Elk08} & \textbf{Liu16} & \textbf{RP}$_{\rho}$ &    \textbf{RP} & \textbf{Nat13} & \textbf{Elk08} & \textbf{Liu16} & \textbf{RP}$_{\rho}$ &    \textbf{RP} & \textbf{Nat13} & \textbf{Elk08} & \textbf{Liu16} & \textbf{RP}$_{\rho}$ &    \textbf{RP} & \textbf{Nat13} & \textbf{Elk08} & \textbf{Liu16} \\
\midrule
\textbf{plane}   & 0.288 &  0.225 & 0.224 &  0.225 &  0.207 &  \textbf{0.550} &  0.261 & 0.235 &  0.225 &  0.217 &  \textbf{0.550} &  0.285 & 0.251 &  0.245 &  0.248 &  \textbf{0.550} &  0.196 & 0.171 &  0.171 &  0.159 &  \textbf{0.550} \\
\textbf{auto} & 0.384 &  0.350 & 0.317 &  0.312 &  0.316 &  \textbf{0.550} &  0.342 & 0.335 &  0.331 &  0.331 &  \textbf{0.550} &  0.328 & 0.348 &  0.334 &  0.333 &  \textbf{0.550} &  0.256 & 0.257 &  0.259 &  0.261 &  \textbf{0.550} \\
\textbf{bird}       & 0.198 &  0.160 & 0.169 &  0.166 &  0.161 &  \textbf{0.550} &  0.188 & 0.185 &  0.179 &  0.177 &  \textbf{0.550} &  0.186 & 0.173 &  0.174 &  0.175 &  \textbf{0.550} &  0.150 & 0.154 &  0.150 &  0.147 &  \textbf{0.550} \\
\textbf{cat}        & 0.188 &  0.164 & 0.175 &  0.174 &  0.175 &  \textbf{0.550} &  0.163 & 0.169 &  0.168 &  0.170 &  \textbf{0.550} &  0.148 & 0.156 &  0.154 &  0.152 &  \textbf{0.550} &  0.145 & 0.143 &  0.140 &  0.145 &  \textbf{0.550} \\
\textbf{deer}       & 0.215 &  0.161 & 0.177 &  0.180 &  0.183 &  \textbf{0.550} &  0.194 & 0.180 &  0.180 &  0.182 &  \textbf{0.550} &  0.174 & 0.175 &  0.176 &  0.175 &  \textbf{0.550} &  0.151 & 0.152 &  0.146 &  0.151 &  \textbf{0.550} \\
\textbf{dog}        & 0.188 &  0.162 & 0.161 &  0.165 &  0.155 &  \textbf{0.550} &  0.175 & 0.160 &  0.161 &  0.158 &  \textbf{0.550} &  0.173 & 0.169 &  0.162 &  0.164 &  \textbf{0.550} &  0.145 & 0.142 &  0.139 &  0.133 &  \textbf{0.550} \\
\textbf{frog}       & 0.318 &  0.246 & 0.264 &  0.262 &  0.258 &  \textbf{0.550} &  0.292 & 0.277 &  0.272 &  0.273 &  \textbf{0.550} &  0.276 & 0.274 &  0.277 &  0.277 &  \textbf{0.550} &  0.239 & 0.212 &  0.206 &  0.212 &  \textbf{0.550} \\
\textbf{horse}      & 0.319 &  0.242 & 0.267 &  0.269 &  0.260 &  \textbf{0.550} &  0.283 & 0.264 &  0.264 &  0.263 &  \textbf{0.550} &  0.288 & 0.282 &  0.279 &  0.278 &  \textbf{0.550} &  0.223 & 0.218 &  0.208 &  0.207 &  \textbf{0.550} \\
\textbf{ship}       & 0.317 &  0.257 & 0.267 &  0.271 &  0.248 &  \textbf{0.550} &  0.296 & 0.266 &  0.267 &  0.259 &  \textbf{0.550} &  0.279 & 0.268 &  0.259 &  0.262 &  \textbf{0.550} &  0.220 & 0.212 &  0.207 &  0.191 &  \textbf{0.550} \\
\textbf{truck}      & 0.329 &  0.288 & 0.261 &  0.271 &  0.263 &  \textbf{0.550} &  0.298 & 0.275 &  0.286 &  0.284 &  \textbf{0.550} &  0.289 & 0.272 &  0.276 &  0.277 &  \textbf{0.550} &  0.241 & 0.213 &  0.208 &  0.204 &  \textbf{0.550} \\
\midrule
\textbf{AVG}      & 0.274 &  0.226 & 0.228 &  0.229 &  0.223 &  \textbf{0.550} &  0.249 & 0.235 &  0.233 &  0.231 &  \textbf{0.550} &  0.243 & 0.237 &  0.234 &  0.234 &  \textbf{0.550} &  0.197 & 0.187 &  0.183 &  0.181 &  \textbf{0.550} \\

\bottomrule
\end{tabular}
}
\end{sc}
\end{small}
\end{center}
\vskip -0.08in
\end{supptable*}


\begin{supptable*}[ht!]

\setlength\tabcolsep{1pt} 
\renewcommand{\arraystretch}{0.85}
\caption{Comparison of \textbf{error} for one-vs-rest CIFAR-10 (averaged over all images) using a \textbf{CNN} classifier. Except for $RP_\rho$, $\rho_1$, $\rho_0$ are given to all methods. Top model scores are in bold with $RP_\rho$ in red if better (smaller) than non-RP models.} 
\vskip -0.1in
\label{table:cifar_cnn_error}
\begin{center}
\begin{small}
\begin{sc}

\resizebox{\textwidth}{!}{
\begin{tabular}{l|c|ccccc|ccccc|ccccc|ccccc}
\toprule

\multicolumn{1}{c}{} &  
\multicolumn{1}{c|}{} & 
\multicolumn{5}{c|}{$\pi_1$\textbf{ = 0}}   &   
\multicolumn{5}{c|}{$\pi_1$\textbf{ = 0.25}}  & 
\multicolumn{10}{c}{$\pi_1$\textbf{ = 0.5}}   \\

\multicolumn{1}{c}{} &  
\multicolumn{1}{c|}{} & 
\multicolumn{5}{c|}{$\rho_1$\textbf{ = 0.5}} & 
\multicolumn{5}{c|}{$\rho_1$\textbf{ = 0.25}}  & 
\multicolumn{5}{c}{$\rho_1$\textbf{ = 0}} & 
\multicolumn{5}{c}{$\rho_1$\textbf{ = 0.5}}    \\

{\textbf{IMAGE}} &   \textbf{True} & \textbf{RP}$_{\rho}$ &    \textbf{RP} & \textbf{Nat13} & \textbf{Elk08} & \textbf{Liu16} & \textbf{RP}$_{\rho}$ &    \textbf{RP} & \textbf{Nat13} & \textbf{Elk08} & \textbf{Liu16} & \textbf{RP}$_{\rho}$ &    \textbf{RP} & \textbf{Nat13} & \textbf{Elk08} & \textbf{Liu16} & \textbf{RP}$_{\rho}$ &    \textbf{RP} & \textbf{Nat13} & \textbf{Elk08} & \textbf{Liu16} \\
\midrule
\textbf{plane}   & 0.044 &  \textcolor{red}{\textbf{0.054}} & \textbf{0.057} &  0.059 &  0.063 &  0.900 &  \textcolor{red}{\textbf{0.050}} & \textbf{0.051} &  0.054 &  0.057 &  0.900 &  \textcolor{red}{\textbf{0.048}} & \textbf{0.045} &  0.049 &  0.048 &  0.100 &  \textcolor{red}{\textbf{0.063}} & \textbf{0.061} &  0.074 &  0.065 &  0.900 \\
\textbf{auto} & 0.021 &  \textcolor{red}{\textbf{0.040}} & \textbf{0.037} &  0.041 &  0.043 &  0.100 &  \textcolor{red}{\textbf{0.032}} & \textbf{0.034} &  0.040 &  0.039 &  0.900 &  0.028 & \textbf{0.026} &  \textbf{0.026} &  \textbf{0.026} &  0.100 &  \textcolor{red}{\textbf{0.047}} & \textbf{0.049} &  0.062 &  0.070 &  0.900 \\
\textbf{bird}       & 0.055 &  0.083 & \textbf{0.078} &  0.080 &  0.082 &  0.900 &  \textcolor{red}{\textbf{0.074}} & \textbf{0.074} &  0.077 &  0.078 &  0.900 &  0.072 & \textbf{0.066} &  0.072 &  0.070 &  0.100 &  0.124 & \textbf{0.084} &  0.089 &  0.093 &  0.900 \\
\textbf{cat}        & 0.077 &  0.108 & \textbf{0.091} &  0.092 &  0.095 &  0.100 &  0.111 & 0.090 &  \textbf{0.086} &  0.089 &  0.900 &  0.113 & \textbf{0.084} &  0.086 &  0.088 &  0.100 &  0.117 & 0.098 &  \textbf{0.094} &  0.100 &  0.900 \\
\textbf{deer}       & 0.049 &  0.081 & \textbf{0.078} &  \textbf{0.078} &  0.079 &  0.900 &  0.080 & \textbf{0.069} &  0.075 &  0.070 &  0.900 &  0.076 & 0.062 &  \textbf{0.061} &  0.062 &  0.100 &  0.106 & \textbf{0.086} &  0.091 &  0.093 &  0.900 \\
\textbf{dog}        & 0.062 &  \textcolor{red}{\textbf{0.075}} & \textbf{0.071} &  0.079 &  0.080 &  0.100 &  0.071 & 0.069 &  0.070 &  \textbf{0.067} &  0.900 &  0.069 & 0.061 &  \textbf{0.057} &  0.076 &  0.100 &  0.103 & \textbf{0.081} &  0.084 &  0.086 &  0.900 \\
\textbf{frog}       & 0.038 &  0.050 & \textbf{0.048} &  \textbf{0.048} &  0.054 &  0.100 &  \textcolor{red}{\textbf{0.047}} & \textbf{0.052} &  0.056 &  0.062 &  0.900 &  0.045 & \textbf{0.040} &  0.042 &  0.043 &  0.100 &  \textcolor{red}{\textbf{0.058}} & \textbf{0.062} &  0.066 &  0.071 &  0.900 \\
\textbf{horse}      & 0.035 &  \textcolor{red}{\textbf{0.050}} & \textbf{0.052} &  0.057 &  0.054 &  0.900 &  \textcolor{red}{\textbf{0.048}} & \textbf{0.051} &  0.052 &  0.057 &  0.900 &  0.045 & \textbf{0.040} &  0.042 &  0.046 &  0.100 &  \textcolor{red}{\textbf{0.065}} & \textbf{0.063} &  0.066 &  0.075 &  0.900 \\
\textbf{ship}       & 0.028 &  \textcolor{red}{\textbf{0.042}} & \textbf{0.042} &  0.046 &  \textbf{0.042} &  0.900 &  \textcolor{red}{\textbf{0.037}} & \textbf{0.036} &  0.042 &  0.047 &  0.900 &  0.035 & 0.033 &  \textbf{0.031} &  0.033 &  0.100 &  \textcolor{red}{\textbf{0.051}} & \textbf{0.049} &  0.064 &  0.058 &  0.900 \\
\textbf{truck}      & 0.027 &  \textcolor{red}{\textbf{0.044}} & \textbf{0.046} &  0.054 &  0.056 &  0.900 &  \textcolor{red}{\textbf{0.034}} & \textbf{0.032} &  0.038 &  0.043 &  0.900 &  \textcolor{red}{\textbf{0.034}} & \textbf{0.031} &  0.034 &  0.034 &  0.100 &  \textcolor{red}{\textbf{0.060}} & 0.066 &  0.067 &  \textbf{0.065} &  0.900 \\
\midrule
\textbf{AVG}      & 0.043 &  \textcolor{red}{\textbf{0.063}} & \textbf{0.060} &  0.064 &  0.065 &  0.580 &  \textcolor{red}{\textbf{0.059}} & \textbf{0.056} &  0.059 &  0.061 &  0.900 &  0.056 & \textbf{0.049} &  0.050 &  0.053 &  0.100 &  0.080 & \textbf{0.070} &  0.076 &  0.077 &  0.900 \\

\bottomrule
\end{tabular}
}
\end{sc}
\end{small}
\end{center}
\vskip -0.08in
\end{supptable*}


\begin{supptable*}[ht!]

\setlength\tabcolsep{1pt} 
\renewcommand{\arraystretch}{0.85}
\caption{Comparison of \textbf{AUC-PR} for one-vs-rest CIFAR-10 (averaged over all images) using a \textbf{CNN} classifier. Except for $RP_\rho$, $\rho_1$, $\rho_0$ are given to all methods. Top model scores are in bold with $RP_\rho$ in red if greater than non-RP models.} 
\vskip -0.1in
\label{table:cifar_cnn_auc}
\begin{center}
\begin{small}
\begin{sc}

\resizebox{\textwidth}{!}{
\begin{tabular}{l|c|ccccc|ccccc|ccccc|ccccc}
\toprule

\multicolumn{1}{c}{} &  
\multicolumn{1}{c|}{} & 
\multicolumn{5}{c|}{$\pi_1$\textbf{ = 0}}   &   
\multicolumn{5}{c|}{$\pi_1$\textbf{ = 0.25}}  & 
\multicolumn{10}{c}{$\pi_1$\textbf{ = 0.5}}   \\

\multicolumn{1}{c}{} &  
\multicolumn{1}{c|}{} & 
\multicolumn{5}{c|}{$\rho_1$\textbf{ = 0.5}} & 
\multicolumn{5}{c|}{$\rho_1$\textbf{ = 0.25}}  & 
\multicolumn{5}{c}{$\rho_1$\textbf{ = 0}} & 
\multicolumn{5}{c}{$\rho_1$\textbf{ = 0.5}}    \\

{\textbf{IMAGE}} &   \textbf{True} & \textbf{RP}$_{\rho}$ &    \textbf{RP} & \textbf{Nat13} & \textbf{Elk08} & \textbf{Liu16} & \textbf{RP}$_{\rho}$ &    \textbf{RP} & \textbf{Nat13} & \textbf{Elk08} & \textbf{Liu16} & \textbf{RP}$_{\rho}$ &    \textbf{RP} & \textbf{Nat13} & \textbf{Elk08} & \textbf{Liu16} & \textbf{RP}$_{\rho}$ &    \textbf{RP} & \textbf{Nat13} & \textbf{Elk08} & \textbf{Liu16} \\
\midrule
\textbf{plane}   & 0.856 &  0.779 & 0.780 &  \textbf{0.784} &  0.756 &  0.550 &  \textcolor{red}{\textbf{0.808}} & \textbf{0.797} &  0.770 &  0.742 &  0.550 &  \textcolor{red}{\textbf{0.813}} & \textbf{0.824} &  0.792 &  0.794 &  0.550 &  \textcolor{red}{\textbf{0.710}} & \textbf{0.722} &  0.662 &  0.682 &  0.550 \\
\textbf{auto} & 0.954 &  0.874 & \textbf{0.889} &  0.878 &  0.833 &  0.550 &  \textcolor{red}{\textbf{0.905}} & \textbf{0.900} &  0.871 &  0.866 &  0.550 &  \textcolor{red}{\textbf{0.931}} & \textbf{0.927} &  0.924 &  0.910 &  0.550 &  \textcolor{red}{\textbf{0.824}} & \textbf{0.814} &  0.756 &  0.702 &  0.550 \\
\textbf{bird}       & 0.761 &  0.559 & 0.566 &  \textbf{0.569} &  0.568 &  0.550 &  \textcolor{red}{\textbf{0.619}} & \textbf{0.618} &  0.584 &  0.597 &  0.550 &  \textcolor{red}{\textbf{0.623}} & \textbf{0.679} &  0.613 &  0.619 &  0.115 &  0.465 & 0.492 &  0.436 &  0.434 &  \textbf{0.550} \\
\textbf{cat}        & 0.601 &  0.387 & 0.447 &  \textbf{0.463} &  0.433 &  0.550 &  0.423 & 0.454 &  0.487 &  0.480 &  \textbf{0.550} &  0.483 & \textbf{0.512} &  0.493 &  0.473 &  0.050 &  0.373 & 0.375 &  0.382 &  0.371 &  \textbf{0.550} \\
\textbf{deer}       & 0.820 &  \textcolor{red}{\textbf{0.620}} & 0.600 &  \textbf{0.615} &  0.573 &  0.550 &  0.646 & \textbf{0.660} &  0.610 &  0.657 &  0.550 &  0.658 & \textbf{0.707} &  0.700 &  0.703 &  0.550 &  0.434 & 0.487 &  0.414 &  0.435 &  \textbf{0.550} \\
\textbf{dog}        & 0.758 &  \textcolor{red}{\textbf{0.629}} & \textbf{0.662} &  0.617 &  0.573 &  0.550 &  \textcolor{red}{\textbf{0.673}} & \textbf{0.667} &  0.658 &  0.660 &  0.550 &  0.705 & 0.722 &  \textbf{0.741} &  0.705 &  0.550 &  0.541 & 0.545 &  0.496 &  0.519 &  \textbf{0.550} \\
\textbf{frog}       & 0.891 &  0.812 & \textbf{0.815} &  0.812 &  0.776 &  0.550 &  \textcolor{red}{\textbf{0.821}} & \textbf{0.827} &  0.808 &  0.749 &  0.550 &  \textcolor{red}{\textbf{0.841}} & \textbf{0.851} &  0.828 &  0.831 &  0.550 &  \textcolor{red}{\textbf{0.753}} & \textbf{0.710} &  0.691 &  0.620 &  0.550 \\
\textbf{horse}      & 0.897 &  \textcolor{red}{\textbf{0.810}} & \textbf{0.817} &  0.799 &  0.779 &  0.550 &  \textcolor{red}{\textbf{0.824}} & \textbf{0.809} &  0.801 &  0.772 &  0.550 &  \textcolor{red}{\textbf{0.826}} & \textbf{0.844} &  0.818 &  0.819 &  0.550 &  \textcolor{red}{\textbf{0.736}} & \textbf{0.699} &  \textbf{0.699} &  0.600 &  0.550 \\
\textbf{ship}       & 0.922 &  \textcolor{red}{\textbf{0.870}} & 0.862 &  \textbf{0.864} &  0.853 &  0.550 &  \textcolor{red}{\textbf{0.889}} & \textbf{0.885} &  0.843 &  0.848 &  0.550 &  0.889 & \textbf{0.897} &  0.891 &  0.887 &  0.550 &  \textcolor{red}{\textbf{0.800}} & \textbf{0.808} &  0.767 &  0.741 &  0.550 \\
\textbf{truck}      & 0.929 &  \textcolor{red}{\textbf{0.845}} & \textbf{0.848} &  0.824 &  0.787 &  0.550 &  \textcolor{red}{\textbf{0.887}} & \textbf{0.894} &  0.873 &  0.853 &  0.550 &  \textcolor{red}{\textbf{0.904}} & \textbf{0.902} &  0.898 &  0.883 &  0.550 &  \textcolor{red}{\textbf{0.740}} & \textbf{0.709} &  0.695 &  0.690 &  0.550 \\
\midrule
\textbf{AVG}      & 0.839 &  0.719 & \textbf{0.729} &  0.722 &  0.693 &  0.550 &  \textcolor{red}{\textbf{0.750}} & \textbf{0.751} &  0.730 &  0.722 &  0.550 &  0.767 & \textbf{0.787} &  0.770 &  0.762 &  0.457 &  \textcolor{red}{\textbf{0.637}} & \textbf{0.636} &  0.600 &  0.579 &  0.550 \\

\bottomrule
\end{tabular}
}
\end{sc}
\end{small}
\end{center}
\end{supptable*}




\end{document}